\documentclass{article}

\usepackage{arxiv}
\UseRawInputEncoding
\usepackage[utf8]{inputenc} 
\usepackage[T1]{fontenc}    
\usepackage{hyperref}       
\usepackage{url}            
\usepackage{booktabs}       
\usepackage{amsfonts}       
\usepackage{nicefrac}       
\usepackage{microtype}      
\usepackage{lipsum}

\usepackage{graphicx}
\usepackage{amssymb}
\usepackage{amsthm}
\usepackage{amsmath}

\usepackage[numbers,sort&compress ]{natbib}

\usepackage[table]{xcolor}
\usepackage{xcolor}

\usepackage{pdflscape}
\usepackage{multirow}

\usepackage[linesnumbered,ruled,vlined]{algorithm2e}
\usepackage{lineno}

\title{Assigning Apples to Individual Trees in Dense Orchards using 3D Color Point Clouds}

\author{
  Mouad Zine-El-Abidine \\
  Laboratoire Angevin de Recherche en Ingénierie des Systèmes (LARIS) \\
   Université d'Angers \\
   Angers, France 
   \And
  Helin Dutagaci \\
  Department of Electrical-Electronics Engineering\\
  Eskisehir Osmangazi University\\
  Eskisehir, Turkey \\
  \texttt{hdutagaci@ogu.edu.tr} \\
   \And
   Gilles Galopin \\
   INRAe, UMR1345 \\
   Institut de Recherche en Horticulture et Semences \\ Angers, France
   \AND
   David Rousseau \\
   Laboratoire Angevin de Recherche en Ingénierie des Systèmes (LARIS) \\
   Université d'Angers \\
   Angers, France \\
   \texttt{david.rousseau@univ-angers.fr}

}

\begin{document}
\maketitle

\begin{abstract}
We propose a 3D color point cloud processing pipeline to count apples on individual apple trees in trellis structured orchards. Fruit counting at the tree level requires separating trees, which is challenging in dense orchards. We employ point clouds acquired from the leaf-off orchard in winter period, where the branch structure is visible, to delineate tree crowns. We localize apples in point clouds acquired in harvest period. Alignment of the two point clouds enables mapping apple locations to the delineated winter cloud and assigning each apple to its bearing tree. Our apple assignment method achieves an accuracy rate higher than 95\%. In addition to presenting a first proof of feasibility, we also provide suggestions for further improvement on our apple assignment pipeline.
\end{abstract}


\section{Introduction}

Apple yield is an important trait for both orchard management and variety testing of apple trees. Manual fruit counting is usually conducted by sampling a fixed percentage (e.g. 5 or 10\%) of trees randomly or systematically and extrapolating the counts on these trees for total yield estimation of the entire orchard \cite{Wulfsohn2012}. This sampling and extrapolation process, in addition to being time-consuming and labor-intensive, does not always produce the desired precision of yield estimation. Computer vision techniques, on the other hand, provide a faster and more accurate alternative to manual counting of fruits \cite{Gongal2015}.

\begin{figure*}[!ht]
\centering
\includegraphics[width=0.95\textwidth]{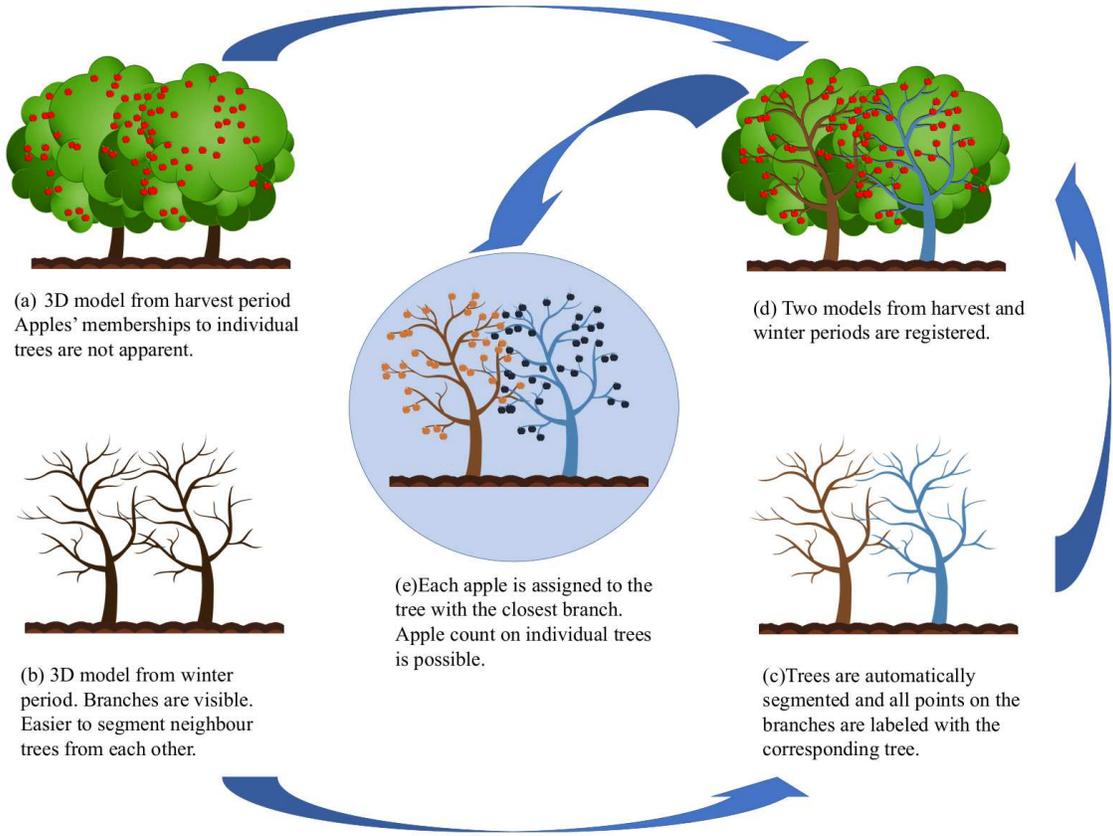}
\caption{Apple detection algorithms usually estimate the cumulative apple count from the harvest season. Our aim is to count the number of apples on each individual tree. The main idea is to register the 3D model from the harvest period (a) with the delineated 3D model from the winter period (c) to align the branches with the detected apples (d). We assign a different label to each delineated tree as an output of the automatic tree separation algorithm we perform on the winter model (c). Finally the detected apples from the harvest model are mapped to their closest branches, and membership of each apple to an individual tree is determined (e).}
\label{fig:intro}
\end{figure*}

While the majority of computer vision techniques for fruit counting relied on RGB (Red, Green, Blue) images, other types of data including RGB-Depth images \cite{GeneMola2019b, Nguyen2016, Tao2017,Tu2018, Lin2019,Fu2020}, spectral images \cite{Safren2007}, thermal images \cite{Stajnko2004, Bulanon2008, Bulanon2009,  Wachs2010, Gan2020} images or LiDAR (Light Detection and Ranging) data \cite{GeneMola2019a} have also been used. In traditional approaches for fruit detection through such sensor information, relevant information is extracted from each data instance separately according to a manually predefined algorithm. The representative quantitative information obtained in this manner is generally referred to as a hand-crafted feature. Hand-crafted approaches can involve techniques such as color thresholding, color space clustering, shape analysis, blob detection, circular Hough transform, Ncut algorithm, employment of Histogram of Oriented Gradients (HOG), Local Binary Patterns (LBP) and Upright Speeded Up Robust Features (U-SURF) for separating fruits from the canopy \cite{Wang2012, SenGupta2014, Sabzi2018, Tao2017, Gongal2016, Nguyen2016, Roy2016a, Bargoti2017b, samiei2020toward, Sun2019, Gong2013, Lu2018, He2020, Kelman2014, Linker2018, Wu2019}. Recently, deep learning methods have become commonplace for fruit detection and counting \cite{ApoloApolo2020, Bargoti2017a, Bresilla2019, Chen2017, Hani2018, HaniIsler2020, Tian2019, GeneMola2019b, Liu2018, Tu2018, Williams2019, Fu2020, Gan2020, Xiong2020}. Deep neural networks are employed to learn predictors from a set of training data through optimizing the parameters of feature extraction and localization of fruits simultaneously. After prediction, further processing, such as circular Hough transform and watershed transform \cite{Bargoti2017b} for verification and filtering of multiple counts through 3D (3-Dimensional) reconstruction \cite{GeneMola2020, Gongal2016, HaniIsler2020} can be applied to extract the final fruit count.

The main objective of most fruit counting methods is to estimate the total number of observable fruits in the sensed data \cite{GeneMola2020, Gongal2016, Hani2018, HaniIsler2020, Liu2018, Bargoti2017a, Bargoti2017b, Bresilla2019, Chen2017, Fu2020}. The fruits are not mapped to their bearing trees; i.e. the number of fruits on each tree is not computed. Examples to applications that will benefit from fruit counting on individual trees are precise yield mapping at tree scale, management of individual trees to maximize uniformity within the orchard, and individual tree-based analysis in variety testing experiments.

Estimation of fruit count on each tree requires separating individual trees and identifying which tree each detected fruit belongs to (\emph{tree membership of the fruit}). Individual tree delineation is the process of separating individual trees, including trunk detection and crown boundary delineation; i.e. identifying the trunk and branches belonging to a single tree \cite{Zhen2016}. Delineation of trees in dense orchards or forests is a challenging task due to interlacing and touching branches of adjacent trees, particularly when there is high variation among the trees in terms of crown size and shape \cite{Zhen2016}. Occlusion caused by dense leaf cover during harvest period further complicates the delineation of trees. Using leaf-off data collected during winter can alleviate the occlusion and facilitate the capture of trunk and branch geometry \cite{Brandtberg2003, Lu2014}. 

The architectural structure that determines the connectivity of the branches to a particular tree trunk becomes ambiguous in 2D images, even during winter period. 2D projection causes loss of shape and connectivity information of the branches of neighboring trees. Processing 3D point clouds is more adequate for our application since 3D data enables a detailed analysis of the geometric structure of trees and localization of branches and fruits in the 3D world. 

Furthermore, acquiring 3D information of the trees in the orchards facilitates a number of applications in precision agriculture, robotic agriculture, and phenotyping. These applications include robotic crop harvesting \cite{Barnea2016, Ge2020, Lin2019, Williams2019}, automated pruning \cite{Medeiros2017, He2018}, monitoring pruning operations \cite{Mendez2016}, and 3D visualization tools to guide the agronomists \cite{YandunNarvaez2016}. Accurate measurements of morphological traits such as canopy volume, branch dimensions and leaf area from already available 3D models are essential for phenotyping experiments and productivity assessment \cite{rosell2012review, coupel2019multi, Tabb2017}.

Computer vision techniques aiding management of fruit orchards range from complete processing pipelines to algorithms performing single tasks such as tree localization \cite{Tabb2017,ColmeneroMartinez2018,Medeiros2017,Zhang2017,Zeng2020,Nielsen2012,underwood2015lidar,zhong2016segmentation,bargoti2015pipeline}. A vision system was developed by \cite{Tabb2017} to reconstruct 3D fruit trees and identify branch structure and traits for automatic pruning. In \cite{ColmeneroMartinez2018} an automatic trunk-detection system using an infrared sensor was introduced. Medeiros et al. \cite{Medeiros2017} employed a laser sensor to model dormant fruit trees and identify primary branches for automatic pruning. In \cite{Zhang2017} Regions-Convolutional Neural Network (R-CNN) was applied on depth images for detection of branches of apple trees and localization of shaking points to guide a harvesting machine. Zeng et al. \cite{Zeng2020} developed an algorithm to segment trellis wires, support poles, and tree trunks in sparse LiDAR point clouds acquired from trellis-structured apple orchards. In order to optimize the mechanization of fruitlet and blossom thinning, Nielsen et al. \cite{Nielsen2012} used LiDAR and stereo vision together for obtaining 3D models of orchard rows of trees. They fitted mixtures of Gaussians to the point cloud to cluster the trees into Gaussian shaped cylinders. In \cite{underwood2015lidar}, LiDAR data was used for individual tree separation through a hidden semi-Markov model. Their objective was to develop a pipeline for building detailed orchard maps and an algorithm to match subsequent LiDAR tree scans to the prior database, enabling correct data association for precision agricultural applications. In \cite{zhong2016segmentation}, a procedure for segmenting canopy to individual trees was proposed. The procedure involved octree construction, clustering, trunk detection and Ncut segmentation. 3D data was obtained with terrestrial laser scanning (TLS) and mobile laser scanning (MLS). In \cite{bargoti2015pipeline}, a tree trunk detection pipeline was proposed for identifying individual trees in a trellis structured apple orchard, using ground-based LiDAR and image data. Hough transformation was performed on 3D point cloud to search for trunk candidates. These candidates were projected into the camera images, where pixel-wise classification was used to update their likelihood of being a tree trunk. Detection was achieved by using a hidden semi-Markov model to leverage from the contextual information provided by the repetitive structure of the orchard.

The objective of this work is to delineate apple trees in a trellis structured orchard and count the number of apples on each individual tree (Fig. \ref{fig:intro}). To the best of our knowledge, this problem was not addressed before in previous works dealing with apple detection and counting. Our strategy is to reconstruct 3D models of the same set of trees twice a year, once during the winter period and once during the harvest period. We perform delineation of individual trees on the leaf-off model from winter, which we refer to as \emph{winter point cloud}. We detect tree trunks and identify the branches connected to them using winter point cloud. We employ the 3D model from the harvest period, which we call \emph{harvest point cloud}, to localize apples. We determine the tree-membership of each apple in the harvest point cloud by mapping their locations onto the winter point cloud, where individual trees are separated. This approach of registering data from two different time instances for fruit counting is another novelty we introduce to the field. We also propose the use of a known calibration object to facilitate the registration of two point clouds and to recover the true metric sizes of the important structures in the scenes.

The main contributions of this study are:
\begin{itemize}
    \item Addressing the problem of apple counting on individual trees from 3D color point clouds.
    \item As a way to map detected apples to individual trees, alignment of harvest point cloud to the winter point cloud, where individual trees are automatically delineated.
    \item A complete pipeline for detecting and removing trellis wires and support poles, detecting tree trunks and delineating crowns of individual trees in winter point clouds. 
    \item The use of a calibration object for correct scaling and alignment of point clouds acquired in different time instances.
\end{itemize}

\section{Materials and Methods}

\begin{figure*}[ht]
\centering
\includegraphics[width=0.8\textwidth]{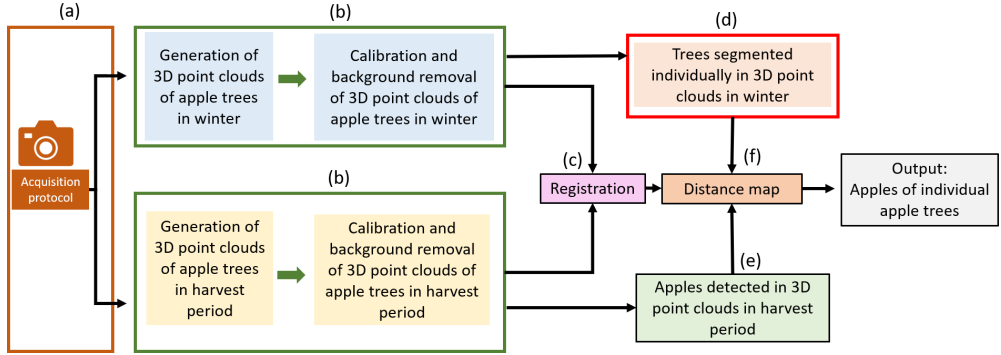}
\caption{Pipeline proposed to assign apples to individual trees. (a) Image acquisition of apple trees in winter and harvest period. (b) Calibration of 3D models and extraction of region of interest. (c) Registration of calibrated models from winter and harvest period. (d)  Separation of individual trees in winter point cloud. (e) Apple detection from harvest point cloud. (f) Distance map to assign apples to individual segmented trees.}
\label{fig:pipeline_full}
\end{figure*}

We developed a point cloud processing pipeline (Fig. \ref{fig:pipeline_full}) in order to locate and count apples on individual trees. We use a color camera for capturing images of target trees in the orchard from multiple views during both winter and harvest periods (Fig. \ref{fig:pipeline_full} (a)). These images are processed by a structure from motion algorithm to reconstruct winter and point clouds. The two point clouds are prepared for initial alignment which we refer to as \emph{calibration of point clouds} (Fig. \ref{fig:pipeline_full} (b)). A novelty of our pipeline is the use of a ColorChecker during acquisition. The ColorChecker serves both as a reference for removal of irrelevant background information and as a calibration tool. Calibration of the point cloud, in our case, involves 1) re-scaling the point cloud to the correct metric scale, 2) orienting the point cloud to a canonical reference frame, 3) extraction of region of interest, and 4) re-centering the point cloud to a predetermined position. The estimated scale allows us to impose metric parameters on the pipeline such as range of separation between trees, separation between trellis wires, diameter of trellis wires, diameter of tree trunks, the expected pole diameter and height, etc. The orientation and re-centering facilitate trellis wire removal, tree trunk detection, and delineation of tree crowns (Fig. \ref{fig:pipeline_full} (d)). The calibration of both harvest and winter point clouds is also crucial for their correct registration (Fig. \ref{fig:pipeline_full} (c)). We employ a color-based apple detection algorithm to locate the apples in the harvest point cloud (Fig. \ref{fig:pipeline_full} (e)). Finally, we map the detected apples onto the winter cloud via distance calculation to assign them to their bearing trees. We give detailed explanations of each module of our pipeline in the following subsections.

\subsection{Experimental Field}

The experiments were conducted in a dense apple orchard, dedicated to variety testing at INRAe-Angers (latitude: 47.48226$^{\circ}$N, longitude: 0.6152$^{\circ}$E) in France. The orchard was composed of 4 years old apple trees organized in I-trellis structure with support poles. Our target trees were arranged in a row, where each tree was a mutant, being tested to be established as a new apple variety. The spacing between trees was 1m in average and the height of the trees ranged from 1 to 3m. The variation of the crown shape among the trees was high.

\subsection{Data acquisition and 3D reconstruction}
\label{sec:scenedata}

\begin{landscape}
\begin{figure}
\centering
\includegraphics[width=1.2\textwidth]{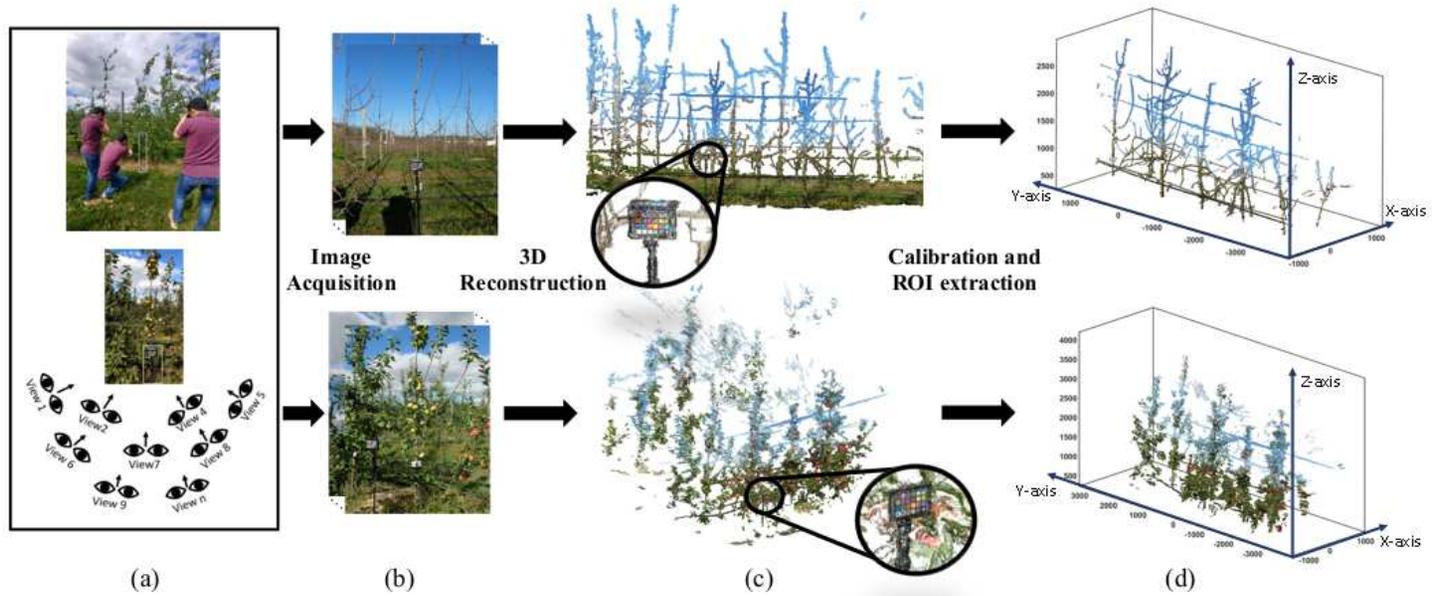}
\caption{Data acquisition and point cloud calibration modules corresponding to (a) and (b) in Fig. \ref{fig:pipeline_full}. (a) Multi-view image acquisition. (b) Apple orchard images acquired in winter and harvest periods. (c) 3D color point cloud reconstructions ($PC_{w}$ and $PC_{h}$) of orchard scenes with zoom on the ColorChecker. (d) 3D color point clouds after calibration and extraction of region of interest ($PC_{w}^{C}$ and $PC_{h}^{C}$). See Supplementary Material A for details of the calibration process.}
\label{fig:Pipeline_data}
\end{figure}
\end{landscape}

Fig. \ref{fig:Pipeline_data} illustrates the data acquisition and point cloud calibration processes of our pipeline, corresponding to the modules (a) and (b) in Fig. \ref{fig:pipeline_full}. We obtained 3D color point clouds of seven scenes from the orchard through a multi-view reconstruction process. A \emph{scene}, in our study, refers to part of an orchard row; i.e. a set of adjacent trees in the same row. Each scene contained 4 to 5 apple trees in our experiments, although our algorithm is capable of processing an entire orchard row. The number of trees in each scene is given in Table \ref{tab:numofimages}. 

A 3D color point cloud (or a 3D RGB point cloud) $PC$ is a set of 3D points, where each point is represented by its coordinates $(x,y,z)$ and its color $(R,G,B)$. Here, $(R,G,B)$ refers to the values of red, green and blue channels.

\begin{table}[!ht]
\centering
\caption{Number of trees in the scenes and number of images acquired in winter and harvest periods.}
\begin{tabular}{{cccc}}
\hline & \small{\#  trees}
& \small{\# images (winter)} & \small{\#  images (harvest)} \\\hline
\small{Scene 1} & \small{5} & \small{236}  & \small{364} \\\hline
\small{Scene 2} & \small{5} & \small{189}  & \small{382} \\\hline
\small{Scene 3} & \small{5} & \small{221}  & \small{380} \\\hline
\small{Scene 4} & \small{4} & \small{183}  & \small{374} \\\hline
\small{Scene 5} & \small{5} & \small{206}  & \small{380} \\\hline
\small{Scene 6} & \small{4} & \small{199}  & \small{376} \\\hline
\small{Scene 7} & \small{4} & \small{227}  & \small{376} \\\hline
\end{tabular}
\label{tab:numofimages} 
\end{table}

We captured multiple RGB images of size $3000 \times 4000$ pixels, of a scene with a color camera (Fujifilm X20, Fujifilm Corporation, Tokyo, Japan) in both winter and harvest periods to reconstruct the point clouds. We acquired images from only one side of the orchard row; although it is possible to follow the procedure proposed in \cite{Roy2018} to reconstruct and register two sides of a row. Table \ref{tab:numofimages} lists the number of images used for 3D reconstruction of the scenes from winter and harvest periods. The viewpoints and viewing angles (i.e. camera positions and orientations) were randomly chosen to get visual information covering the entire scene. In this study, we captured the images manually; however, this process can also be performed by a land robot equipped with multiple cameras. 

The multi-view images were used to reconstruct 3D color point clouds of the scenes through VisualSFM \cite{VSFM1,VSFM2} and PMVS/CMVS tool \cite{PMVS1,PMVS2}. VisualSFM is a freely available software \cite{VSFM1,VSFM2} that performs Structure from Motion (SfM) to estimate unknown camera locations and orientations. It provides a sparse point cloud of the scene through keypoint matching and triangulation. In order to obtain a dense point cloud, we used PMVS/CMVS tool, another freely-available software \cite{PMVS1,PMVS2}. This tool takes as input the images and the camera parameters computed by VisualSFM and provides a dense reconstruction of the scene through multi-view stereo. For introductory and in-depth information on the techniques of SfM and multi-view stereo, we refer the reader to the textbook of Hartley and Zisserman \cite{Hartley2004}.

Before capturing the images of each scene, we installed a calibration object (ColorChecker Passport Photo 2, X-rite, Great Lakes, Midwestern US) mounted on a tripod stick at a known position. We placed the tripod stick in front of the trees facing the camera, such that the ColorChecker pattern is almost parallel to the tree row Fig. \ref{fig:pipeline_full} (b). When the ColorChecker stick was installed, we manually measured two distances with a tape measure: $d_R^{cc}$: the minimum distance of the tripod stick to the tree row, and $d_T^{cc}$: the distance to a designated target tree. These values are necessary for the calibration process of the point clouds.

The reconstructed harvest point cloud and winter point cloud of a scene are referred to as $PC_{h}$ and $PC_{w}$ respectively. Point clouds of a sample scene are given in Fig. \ref{fig:Pipeline_data} (c) with the ColorChecker objects zoomed in.

\subsection{Calibration and Extraction of Region of Interest}

The calibration of the point clouds from harvest and winter periods provides an initial alignment, which is fundamental for the success of the registration of the two point clouds. Having the point cloud with the accurate scale also enables us to fix parameters, such as trunk diameter, tree height, separation between trees, according to the range of expected metric sizes of the structures in the scene.

The ColorChecker is usually employed as a color reference to obtain accurate colors from images under varying lighting conditions \cite{Fernandez2019}. In this work, we do not use the ColorChecker for this purpose. Instead, we use it as a distinct reference pattern to geometrically calibrate the raw point clouds. We developed an algorithm for automatic detection of the ColorChecker, together with the tripod stick it is mounted on, from 3D color point clouds. The description of this algorithm can be found in Supplementary Material A. The 3D locations of the centers of the color patches of the ColorChecker chart are used to guide the calibration of the point cloud. 

The geometric calibration process takes as input the harvest and winter point clouds ($PC_{h}$ and $PC_{w}$) and produces the calibrated point clouds ($PC_{h}^{C}$ and $PC_{w}^{C}$), as shown in Fig. \ref{fig:Pipeline_data} (d). The details of the calibration process are given in Supplementary Material A. In summary, the calibration process consists of 1) estimation of the true scale and re-scaling the point cloud; 2) re-defining a canonical reference frame and rotating the point cloud to this new frame; 3) extraction of region of interest, which corresponds to the set of trees just behind the ColorChecker; and 4) moving the origin of the reference frame to the base of the designated tree. The canonical reference frame is defined such that Y-axis is parallel to the tree row and Z-axis is orthogonal to the ground.


\subsection{Separation of Individual Trees}

In this section, we describe the procedure to separate the trees from each other in the winter scenes. This procedure involves localization of target tree trunks, finding the points on the tree trunks, detecting and removing trellis wires, the water pipe, and the support poles. After the trees are localized and irrelevant points are removed, the tree membership of all the remaining points are determined.

Let the number of points in the calibrated winter point cloud $PC_{w}^{C} = \{p_1,p_2,...,p_{N_W}\}$ be $N_W$. We aim to map each point $p_i$ to a semantic label $\gamma_i$, $i = {1,2,...,N_W}$ where $\gamma_i \in \Gamma$. $\Gamma$ is the set of four semantic labels: $\Gamma =  \{$ \textit{"Tree trunk"}, \textit{"Branch"}, \textit{"Trellis wire+Water pipe"}, \textit{"Support pole"}$\}$. The process of automatically labeling the points in the cloud with one of these four classes is called \emph{semantic segmentation} of the scene. The rationale for a semantic segmentation stage is to remove irrelevant structures and to eliminate the connectivity between adjacent trees caused by trellis wires and the water pipe.

In conjunction with semantic segmentation, we also detect trees in the scene and locate their trunks. Let the set of verified trees in the scene be denoted as $\mathcal{T}$. Each tree $T_j$ in $\mathcal{T}$ is represented by its tree identity $t_j \in \{1,2,...,N_{trees}\}$ and its location $L_j$, for $j = 1,2,...,N_{trees}$. The location of a tree corresponds to the coordinates of its base $L_j = (x_j,y_j,z_j)$, $j = 1,2,...,N_{trees}$ measured in the canonical reference frame.

After removing the irrelevant structures (trellis wires, water pipe and support pole) we delineate the trees in the winter point cloud. The final output of the tree separation algorithm is the assignment of each trunk and branch point in the calibrated winter point cloud $PC_{w}^{C}$ to one of the trees in the set $\mathcal{T}$.

\subsubsection{Detection of trellis wires and tree trunks}
\label{sec:trellis}

\begin{figure*}[th]
\centering
\includegraphics[width=0.79\textwidth]{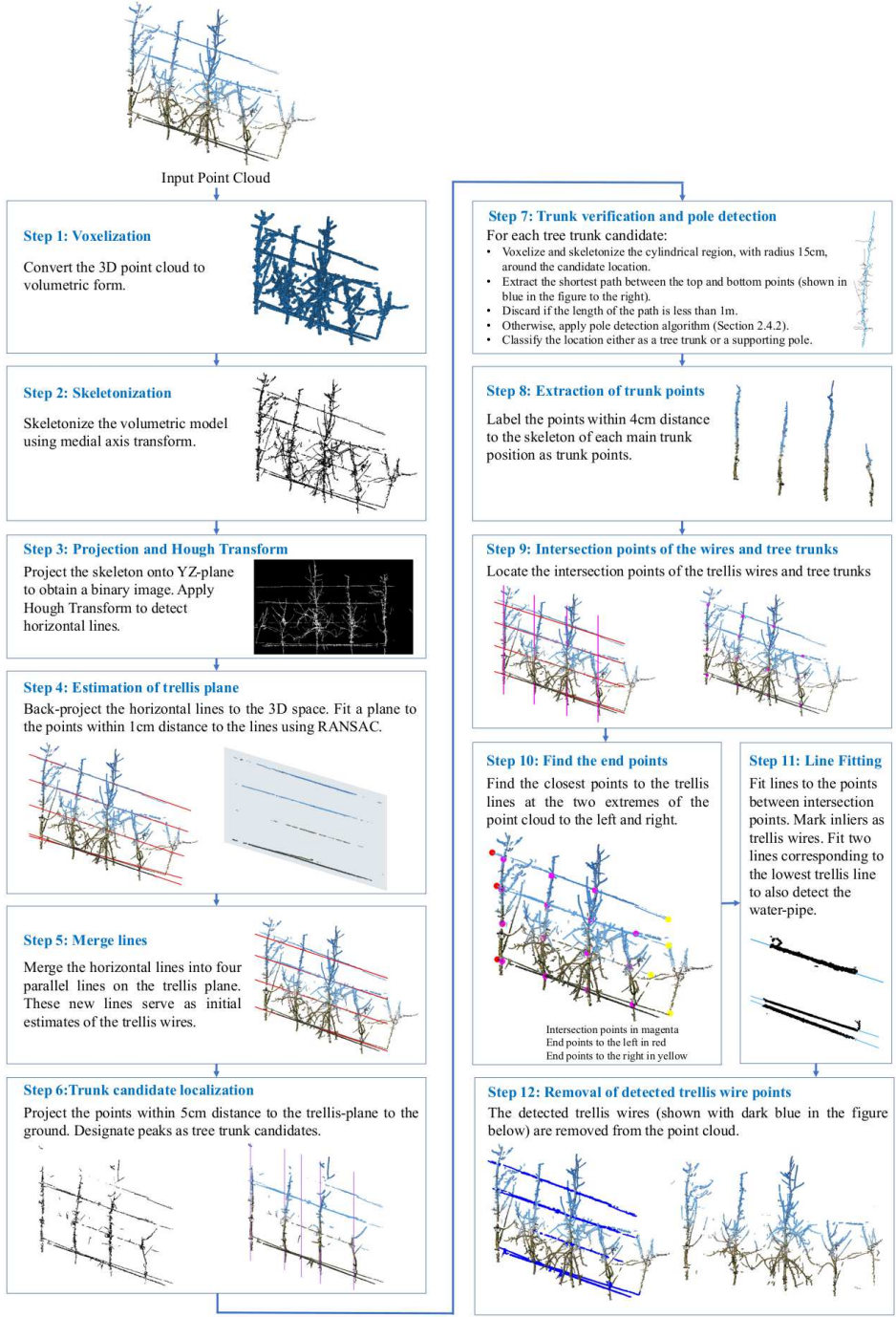}
\caption{Block diagram for detection and removal of trellis wires and the water-pipe.}
\label{fig:blocktrellis}
\end{figure*}

The procedure for detecting points on trellis wires is based-on estimation of the \emph{trellis-plane} and the \emph{trellis-lines} along the trellis wires and operating on the points close to these estimates. Candidate trunk locations are detected along the trellis-plane based on point density. The points in a cylindrical region along each candidate location is separately skeletonized. The skeleton and the points surrounding it are examined to verify tree trunk position and to detect the presence of a supporting pole. 3D points belonging to the trunk of each individual tree and support pole are identified and labeled. Regions between tree trunks along the initial line estimates are re-examined through 3D line fitting to increase the precision of the detection and removal of the points that belong to the trellis wires. The steps of the procedure are shown in Fig. \ref{fig:blocktrellis} and detailed below: 

\underline{\textit{Step 1: Voxelization}} The calibrated winter point cloud $PC_{w}^{C}$ is converted to binary volumetric form, where a voxel takes the value 1 if the voxel is occupied by the points in $PC_{w}^{C}$. Specifically, we fit a regular 3D grid to the bounding box defined by the minimum and maximum coordinate values $(x_{min},x_{max})$, $(y_{min},y_{max})$, $(z_{min},z_{max})$ of the points in $PC_{w}^{C}$. Each cell, i.e. voxel, of the grid has edge lengths of $\Delta_x = \Delta_y =  \Delta_z = 5mm$. On this grid, we define a 3D array $B$ of size $N_x \times N_y\times N_z$, where

\begin{equation}
N_x = \lfloor \frac{x_{max}-x_{min}}{\Delta_x} \rfloor + 1 ; \quad
 N_y = \lfloor \frac{y_{max}-y_{min}}{\Delta_y} \rfloor + 1  ; \quad
  N_z = \lfloor \frac{z_{max}-z_{min}}{\Delta_z} \rfloor + 1 .
\end{equation}

\noindent Here $\lfloor \cdot \rfloor$ is the floor function. The 3D volumetric form of the point cloud corresponds to the binary function $B$ computed as 

\begin{equation}
B(k,l,m) = 
\begin{cases}
    1, & \text{if } \exists p = (x,y,z) \in PC_{w}^{C} \text{ : } \\ & \lfloor \frac{x-x_{min}}{\Delta_x}\rfloor = k \text{ } \& \text{ } \lfloor \frac{y-y_{min}}{\Delta_y}\rfloor = l \text{ }  \& \text{ }  \lfloor \frac{z-z_{min}}{\Delta_z}\rfloor = m \\
    0,              & \text{otherwise},
\end{cases}
\end{equation}

\noindent for $k = 0,...,N_x-1$, $l = 0,...,N_y-1$, and $m = 0,...,N_z-1$. In Fig. \ref{fig:blocktrellis} (Step 1), the volumetric model of a sample scene is visualized. In the figure only the voxels with value "1" are shown.

\clearpage

\underline{\textit{Step 2: Skeletonization}} We extract the skeleton of the volumetric model $B$ using medial axis thinning algorithm given in \cite{skeleton}. Formally, the skeleton of a 3D object is the set of the centers of all inscribed maximal spheres where these spheres touch the object boundary at one than more point \cite{skeleton}. The skeletonization process produces another binary 3D grid $S$ of size $N_x \times N_y\times N_z$, where the structures in $B$ are pruned to curves with thickness of one voxel. In Fig. \ref{fig:blocktrellis} (Step 2), the skeleton of a sample scene is shown.

\underline{\textit{Step 3: Projection and Hough Transform}} The skeleton defined in the binary 3D grid $S$ is projected to the YZ-plane (parallel to the tree row) as a binary image, $IH$ of size $N_y\times N_z$:
\begin{equation}
    IH(l,m) =
    \begin{cases} 
     1, & \text{if  }  \sum_{k=0}^{N_x-1} S(k,l,m) > 0   \\
     0, & \text{otherwise},
    \end{cases}
\end{equation}
\noindent for $l = 0,...,N_y-1$, and $m = 0,...,N_z-1$.

In Fig. \ref{fig:blocktrellis} (Step 3), the projected binary image of a sample scene is shown. We apply 2D Hough Transform \cite{Duda1972} to $IH$ to extract main horizontal lines in the binary image. The peaks greater than 20\% of the maximum value in the Hough parameter space, and with angle with the horizontal axis less than 10$^{\circ}$ are selected as the main horizontal lines. These horizontal lines correspond to candidates for the trellis-lines in the scene. 

\underline{\textit{Step 4: Estimation of the trellis-plane}}

The detected horizontal lines are back-projected to the 3D space of the point cloud $PC_{w}^{C}$, as shown with red lines in Fig. \ref{fig:blocktrellis} (Step 4). Let the set of these horizontal 3D lines be $\mathcal{L}_{HL}=\{hl_1,hl_2,...,hl_{N_{HL}} \}$, where $N_{HL}$ is the number of horizontal lines. Each 3D line is defined by a pair of points on it, as $hl_r =(p_{r,1},p_{r,2})$, with $p_{r,1}=(x_{r,1},y_{r,1},z_{r,1})$ and $p_{r,2}=(x_{r,2},y_{r,2},z_{r,2})$. We retrieve the points in $PC_{w}^{C}$ with distance 1cm to these lines, and form the subset:
\begin{equation}
    PC_{tr} = \{ p = (x,y,z) \in PC_{w}^{C} \text{ : } \min_{r=1,..,N_{HL}}d(p,hl_r) < 1cm \}.
\end{equation}

\noindent The distance $d(p,hl_r)$ between a point $p$ and the line $hl_r$ is calculated as:
\begin{equation}
    d(p,hl_r)=\frac{\|(p-p_{r,1}) \times (p-p_{r,2})\|}{\|p_{r,2}-p_{r,1}\|},
\end{equation}
\noindent where $\times$ is the cross product operation, and $\|\cdot\|$ is the Euclidean norm. We fit a plane to the points in $PC_{tr}$ using M-estimator SAmple Consensus (MSAC) algorithm given in \cite{Torr2000}, which is a variant of RANdom SAmple Consensus (RANSAC) algorithm. Maximum distance for a point to be an inlier is set to be 0.5cm. The output of the algorithm is a plane model $(A,B,C,D)$, where the parameters define the plane equation $Ax+By+Cz+D=0$. The unit vector $n_{TP} = (A,B,C)$ corresponds to the normal of the plane. We refer to this plane as the \emph{trellis-plane} on which trellis wires and tree trunks are located. Fig. \ref{fig:blocktrellis} (Step 4) shows the trellis-plane fitted to the points in $PC_{tr}$ for a sample scene.

The trellis-plane plays an important role in the following steps. We rotate the calibrated winter point cloud $PC_{w}^{C}$ to a new reference frame such that the new YZ plane coincides with the trellis-plane and Y-axis is parallel to the trellis-lines. The new Y-axis is computed as the average of the direction vectors of the horizontal lines in $\mathcal{L}_{HL}$:
\begin{equation}
u_Y = \frac{\sum_{r=1}^{N_{HL}}(p_{r,2}-p_{r,1}) }{\|\sum_{r=1}^{N_{HL}}(p_{r,2}-p_{r,1}) \|}
\end{equation}

\noindent The new Z-axis is orthogonal to the normal of the trellis-plane and the average direction of the trellis-lines:
\begin{equation}
u_Z = u_Y \times  n_{TP},
\end{equation}

\noindent and the new X-axis is 
\begin{equation}
u_X = u_Y \times u_Z
\end{equation}

We transform each point $p=(x,y,z)$ in the calibrated winter cloud $PC_{w}^{C}$ using the rotation matrix $R$ defined in Eq. (\ref{eq:rottotP}), and obtain a point cloud of the same size, $PC_{w}^{TP}$. We refer to this point cloud as the winter point cloud aligned to the trellis-plane.
\begin{equation} \label{eq:rottotP}
    PC_{w}^{TP} = \{ \hat{p}=(\hat{x},\hat{y},\hat{z})=pR \text{ : } p  \in PC_{w}^{C}\}; \quad R = \begin{bmatrix}
    u_X \\
    u_Y \\
    u_Z 
    \end{bmatrix}
\end{equation}

The origin of the new reference frame remains at the base of the target tree (see Supplementary Material A). With the transformation, the trellis-plane coincides with the $\hat{x}=0$ plane in the new reference frame. This ensures that the $\hat{x}$ coordinate of each tree trunk is close to 0. Notice that this transformation is applied only to the winter point cloud. Once the semantic segmentation of the winter cloud is achieved and the trees are delineated, the points are transformed back to their original positions using $p = \hat{p}R^{-1}$.

\underline{\textit{Step 5: Merge lines}}
The detected horizontal lines are on the trellis-plane; hence, they are located on the $\hat{x}=0$ plane in the new reference frame. Their average direction is parallel to the Y-axis. Hence, we represent each line $hl_r \in \mathcal{L}_{HL}$ with the direction vector $(0,1,0)$ and a point on the line $(0,0,\hat{z}_r)$. The value $\hat{z_r}$ indicates the height of a horizontal line on the trellis-plane and is calculated as:
\begin{equation}
    \hat{p}_{r,1}= (\hat{x}_{r,1},\hat{y}_{r,1}, \hat{z}_{r,1}) = p_{r,1}R; \quad \hat{p}_{r,2}= (\hat{x}_{r,2},\hat{y}_{r,2}, \hat{z}_{r,2}) = p_{r,2}R;
\end{equation}
\begin{equation}
     \hat{z}_{r} = \frac{\hat{z}_{r_1}+\hat{z}_{r_2}}{2}
\end{equation}

We merge the lines into parallel lines on the trellis-plane, each separated by at least 30cm to create the set of trellis-lines $\mathcal{L}_{TL}= \{tl_{1},...,tl_{N_{TL}}\}$. Each line is represented with the direction vector $(0,1,0)$ and a point on the line $(0,0,\hat{z}_q)$, with $q=1,...,N_{TL}$. We use the following procedure to cluster the horizontal lines in $\mathcal{L}_{HL}$ into trellis-lines in $\mathcal{L}_{TL}$: We first sort the horizontal lines with ascending height. We start from the bottom line on the trellis-plane, and initialize $\hat{z}_1$ to the height of the first horizontal line. If the distance between the closest horizontal line is less than 30cm, we add the line to the group and update $\hat{z}_1$ to the average height of the group. Otherwise, we create a new group and proceed to the next line.
In our experiments, the horizontal lines were grouped into 4 lines for all the winter scenes. Fig. \ref{fig:blocktrellis} (Step 5) shows the resulting trellis-lines in red color for a sample winter scene. In the rest of the paper we fix $N_{TL}=4$. The four height values $\{  \hat{z}_1, \hat{z}_2, \hat{z}_3, \hat{z}_4 \}$ will be used to specify the locations of the trellis-lines.

\underline{\textit{Step 6: Trunk candidate localization}}
To localize candidate tree trunks along the trellis-plane we limit the search space within 5cm distance to the trellis-plane. We extract a subset of points $PC_{ts}$ from 
$PC_{w}^{TP}$:
\begin{equation}
    PC_{ts} = \{ \hat{p}=(\hat{x},\hat{y},\hat{z})\in PC_{w}^{TP} \text{ : } |\hat{x}| < 5cm\}
\end{equation}

Fig. \ref{fig:blocktrellis} (Step 6) shows $PC_{ts}$ of a sample scene. We define a regular 2D grid, $IG$ on the $z=0$ plane, which is parallel to the ground. Each cell of the grid has edge length $\hat{\Delta}_x=\hat{\Delta}_y=1cm$. We compute the number of points in $PC_{ts}$ falling into each cell:
\begin{equation}
    \mathcal{IG}_{i,j} = \{ \hat{p}=(\hat{x},\hat{y},\hat{z})\in PC_{ts} \text{ : } \lfloor \frac{\hat{x}-\hat{x}_{min}}{\hat{\Delta}_x}\rfloor = i \quad \& \quad \lfloor \frac{\hat{y}-\hat{y}_{min}}{\hat{\Delta}_y}\rfloor = j   \};
\end{equation}
\begin{equation}
    IG(i,j) = |\mathcal{IG}_{i,j}|,
\end{equation}

\noindent where $\hat{x}_{min}$ and $\hat{y}_{min}$ are the minimum of the $\hat{x}$ and $\hat{y}$ coordinates of the points in $PC_{ts}$, and  $|\mathcal{X}|$ denotes the number of elements in the set $\mathcal{X}$.

$IG$ is the histogram of the points in $PC_{ts}$ projected to the ground. The points on the tree trunks form the densest regions in the histogram correspond to the peaks of $IG$. The locations of the peaks are detected via non-maximum suppression \citep{Gonzalez} as $\{(I_1,J_1),(I_2,J_2),...,(I_{N_P},J_{N_P})\}$, where $N_P$ is the number of detected peaks. The set of candidate trunk locations in the 3D space are then defined as $\mathcal{CT}=\{(0,\hat{y}_1^{ct},0), (0,\hat{y}_2^{ct},0),...,(0,\hat{y}_{N_P}^{ct},0)\}$; with $\hat{y}_1 < \hat{y}_1 < ... < \hat{y}_{N_P}^{ct}$. Recall that the trunks intersect with the trellis-plane. $\hat{y}_s^{ct}$ for $s=1,...,N_P$ is calculated as:
\begin{equation}
    \hat{y}_s^{ct} = J_s\hat{\Delta}_y +\hat{y}_{min}. 
\end{equation}

Fig. \ref{fig:blocktrellis} (Step 6) shows the locations of the candidate trunks as vertical purple lines passing through $(0, \hat{y}_s^{ct} ,0)$.

\underline{\textit{Step 7: Trunk verification}}

Not all the peaks detected in the previous step correspond to tree trunks. In this step, we examine the points at each candidate trunk location to verify whether it is a tree trunk, a support pole, or neither. We construct the set of trees $\mathcal{T}$ using the verified trunks. Each tree $T_j$ in $\mathcal{T}$ is represented by its tree identity $t_j \in \{1,2,...,N_{trees}\}$ and the location of its base $\hat{L}_j = (\hat{x}_j,\hat{y}_j,\hat{z}_j)$, for $j = 1,2,...,T_{N_{trees}}$. The procedure for constructing the set of detected trees is given in Algorithm \ref{alg:formT}, and explained below:

We first initialize the set of trees as $\mathcal{T}=\emptyset$ and the number of tree trunks as $N_{trees} = 0$. For each candidate trunk indexed with $s$, we define a cylindrical region, with radius 15 cm, centered at the candidate trunk location $(0,\hat{y}_s^{ct},0)$, along the trellis-plane. We extract the points inside this region from $PC_{w}^{TP}$:
\begin{equation}
    PC_{s}^{CT} = \{ \hat{p}=(\hat{x},\hat{y},\hat{z})\in PC_{w}^{TP} \text{ : } \sqrt{\hat{x}^2+(\hat{y}^2-\hat{y}_s^{ct})^2} < 15cm\}
    \label{eq:truncyln}
\end{equation}
The point cloud $PC_{s}^{CT}$ is converted to binary volumetric form $B_{s}$ with voxel size $\hat{\Delta}_x = \hat{\Delta}_y =  \hat{\Delta}_z = 5mm$. Then, the skeleton $S_{s}$ is extracted from $B_{s}$ using medial axis thinning algorithm given in \cite{skeleton}. The points on the skeleton are retrieved from the point cloud $PC_{s}^{CT}$, and denoted as $SK_{s}$.

\begin{center}
\begin{algorithm}[H]

 \KwData{$PC_{w}^{TP}$: The winter point cloud aligned to the trellis-plane; \\ 
 $(0,\hat{y}_s^{ct},0)$: Candidate tree trunk locations for $s=1,...,N_P$}
 
\KwResult{$\mathcal{T} = \{T_1,...,T_{N_{trees}}\}$: Set of detected trees;\\
$N_{trees}$: Number of detected trees;\\
$t_j$: Tree identity of $T_j \in \mathcal{T}$;\\
$\hat{L}_j = (\hat{x}_j,\hat{y}_j,\hat{z}_j)$: Location of $T_j \in \mathcal{T}$;\\
$SP_j$: Set of points on the main axis of $T_j$
}

 Initialize $\mathcal{T}=\emptyset$; $N_{trees}=0$; $j=0$;

 \For{$s \gets 1$ to $N_P$}{
 
 Extract the point set $PC_s^{CT}$ using Eq. (\ref{eq:truncyln})\;
 Convert $PC_s^{CT}$ to binary volumetric form $B_s$ through voxelization\;
 Compute the skeleton $S_s$ of $B_s$ using medial axis thinning \cite{skeleton}\;
 Obtain $SK_s$ by retrieving the 3D points on the skeleton $S_s$ \;
 Find the top and bottom points in $SK_s$ with the largest and smallest z-coordinates and designate them as $\hat{p}_{s,{top}}$ and $\hat{p}_{s,{bottom}}$ \; 
 Extract the shortest path between $\hat{p}_{s,{top}}$ and $\hat{p}_{s,{bottom}}$ using Breadth-first search \cite{Algorithms} \;
 Collect the points on the shortest path to form the main axis $SP_s$\;
 Calculate the length $d_s^{SP}$ of $SP_s$\; 
 
 \If{$d_s^{SP}>1m$}{
 Run Support Pole Detection Algorithm on $s$ (Section \ref{sec:poles}) \;
    \If{$s$ is not a Support Pole}{
    $j\leftarrow j+1$ \;  $N_{trees} \leftarrow N_{trees}+1$ \; 
    $t_j = j$ \; $\hat{L}_j = (0,\hat{y}_s^{ct},0)$ \; 
    $SP_j = SP_s$ \;
    $T_j = (t_j,\hat{L}_j,SP_j)$ \;
    $\mathcal{T} \leftarrow \mathcal{T} \cup T_j$
 }
 }\
  }
 
 \caption{Tree trunk verification}
 \label{alg:formT}
\end{algorithm}
\end{center}

The  two top and bottom points of the set $SK_{s}$ along the Z-axis $\hat{p}_{s,{top}}$ and $\hat{p}_{s,{bottom}}$ are retrieved. The points on the shortest path between these two points is computed using the Breadth-first search algorithm described in \cite{Algorithms}. We refer to the set of the points on the shortest path as the \emph{main axis} of the $s$\textsuperscript{th} trunk, and denote it as $SP_{s}$. Fig. \ref{fig:blocktrellis} (Step 7) shows the skeleton with black dots and the points on the shortest path with blue dots for a candidate trunk location.

If the length of the shortest path $d_s^{SP}$ is less than 1m, then the candidate trunk location is discarded. Otherwise, it is passed to the support pole detection procedure described in Section \ref{sec:poles}. If it is not identified as a support pole, then we update $N_{trees} \leftarrow N_{trees}+1$, and insert the verified trunk into $\mathcal{T}$. We also store the main axis of the verified trunk. See Algorithm \ref{alg:formT} for the formation of the set $\mathcal{T}$.

\underline{\textit{Step 8: Extraction of trunk points}}
The previous step gives the attributes of each tree $T_j = (t_j,\hat{L}_j,SP_j) \in \mathcal{T}$. The main axis of the $j$\textsuperscript{th} detected tree is represented by the set of points $SP_j$. We label a point $\hat{p}_i$ in the point cloud $PC_w^{TP}$ as \textit{"Tree trunk"} if its distance to the main axis of one of the trees is less than 3cm. Specifically:
\begin{equation}
    \gamma_i = \text{\textit{"Tree trunk"}} \quad \text{ if } \quad \min_{j}\min_{\hat{p} \in {SP_j}} \|\hat{p}-\hat{p}_i\|^2 < 3cm
\end{equation}

Fig. \ref{fig:blocktrellis} (Step 8) shows the points semantically labeled as \textit{"Tree trunk"} in a winter scene.

\underline{\textit{Step 9: Locating the intersection points of trellis wires and tree trunks}} In Step 4, the set of trellis-lines $\mathcal{L}_{TL}= \{tl_{1},tl_{2},tl_{3},tl_{4}\}$ is determined. Recall that each line is represented with the direction vector $(0,1,0)$ and a point on the line $(0,0,\hat{z}_q)$, with $\hat{z}_1 < \hat{z}_2 < \hat{z}_3 < \hat{z}_4  $. Now, having located the trunks at $\hat{L}_j = (0,\hat{y}_j,0)$ with $\hat{y}_1 < \hat{y}_2 < ...  <\hat{y}_{N_{trees}}$, we find the points where the trellis wires intersect with the trunk locations. For a trellis-line with index $q$ and a trunk location with index $j$, we find the point $\hat{p}_{q,j} \in PC_{w}^{TP}$ closest to the location $(0,\hat{y}_j,\hat{z}_q)$. Fig. \ref{fig:blocktrellis} (Step 9) shows the trellis-lines, located tree trunks and the intersection points $\hat{p}_{q,j}$ for a winter scene.

\underline{\textit{Step 10: Finding the end points of the trellis wires}} The end-points corresponding to the trellis wires in the scene are determined by finding the closest points to the trellis-lines at the two extremes of the point cloud along the Y-axis. Specifically, for a trellis-line with index $q$, we locate two points $\hat{p}_{q,0} \in PC_{w}^{TP}$ and $\hat{p}_{q,{N_{trees}}+1}  \in PC_{w}^{TP}$, which are closest to the locations $(0,\hat{y}_{min},z_q)$ and $(0,\hat{y}_{max},z_q)$, respectively. Here, $\hat{y}_{min}$ and $\hat{y}_{max}$ are the minimum and maximum Y-coordinates of the points in $PC_{w}^{TP}$.  Fig. \ref{fig:blocktrellis} (Step 10) shows the end points for a winter scene with red and yellow dots.

\underline{\textit{Step 11: Line fitting to find the points on the trellis wires and the water-pipe}} The region between each adjacent intersecting points of trellis-lines and the trunks are examined for a precise determination of the points on the trellis wires and the water-pipe. For each pair of intersecting points $\hat{p}_{q,j}$ and $\hat{p}_{q,j+1}$, $q=1,...,4   \quad j = 0,1,...,N_{trees}$ we extract the points:
\begin{equation}
\begin{split}
    PC_{j,j+1}^{q} = &
     \{ \hat{p}=(\hat{x},\hat{y},\hat{z})\in PC_{w}^{TP} \text{ : }  \\
     & (\hat{y}_j+4cm<\hat{y}<\hat{y}_{j+1}-4cm)\text{  } \& \text{  } (d(\hat{p},ls_{j,j+1}) < 10cm) \} 
\end{split}
\end{equation}

\noindent where $ls_{j,j+1}$ is the line defined by the points $\hat{p}_{q,j}$ and $\hat{p}_{q,j+1}$, and $d(\hat{p},ls_{j,j+1})$ is the distance between point $\hat{p}$ and line $ls_{j,j+1}$. This region corresponds to a cylinder of radius 10cm with axis $ls_{j,j+1}$. We set an offset value of 4cm from the trunk locations not to include trunk points to the search region for trellis wire points. 

Using MSAC algorithm given in \cite{Torr2000}, we fit two lines to the points in $PC_{j,j+1}^{0}$ corresponding to the regions along the lowest trellis-line (one for the trellis wire and one for the water-pipe). One line is fitted to the points for the rest of the regions $PC_{j,j+1}^{q}$ with $q=2,3,4$. For a point to be an inlier, the maximum distance to the fitted line is set to be 7cm for $q=0$ and 4cm for $q=2,3,4$. Fig. \ref{fig:blocktrellis} (Step 11) shows points in two regions along the trellis wires in blue and the lines fitted to them in black.

If a point $\hat{p}_i \in PC_{w}^{TP}$ is an inlier of one of the fitted lines we set its semantic label as $\gamma_i  = \text{\textit{"Trellis wire + Water pipe"}}$.

\underline{\textit{Step 12: Removal of detected trellis wire points}} 

The detected trellis wire points and the points on the support pole, if there is any, are removed from the point cloud to form the set:
\begin{equation}
\begin{split}
PC_{w}^{trees} = & \{\hat{p_i} \in PC_{w}^{TP} \text{ : } \\ & (\gamma_i \neq \text{\textit{"Trellis wire + Water pipe"}}) \text{  } \& \text{  } (\gamma_i \neq \text{\textit{Support pole"}}) \}
\end{split}
\end{equation}

The procedure for retrieving points on the support pole is given in Section \ref{sec:poles}. The point cloud $PC_{w}^{trees}$ is supposed to include only the points on the trees. In Fig. \ref{fig:blocktrellis}
(Step 12) the points labeled as \textit{"Trellis wire+Water pipe"} are shown in dark blue. Also the resulting $PC_{w}^{trees}$ is given for a sample winter scene.

\subsubsection{Detection of Support Poles}
\label{sec:poles}
During the procedure for trellis wire detection and localization of tree trunks, we examine each trunk candidate to determine whether it corresponds to a support pole or an actual tree trunk. We consider the points in a vertical cylindrical region of radius 15cm centered at the candidate trunk location. We partition the points into horizontal slices of height 2cm. We project the points in each slice onto the XY-plane (the ground plane) and fit a circle of radius 4.5cm (the actual radius of a support pole in the orchard) to the projected points, and estimate the center. The centers of the slices form the axis of the candidate support pole and the new cylindrical region. We count the points in the cylindrical shell with inner and outer radii, $4.5-0.5$ and $4.5+0.5$cm, and with height 2.3m (the actual height of a support pole). If the ratio of this number to the total number of points in the initial cylindrical region is higher than 0.8, then we declare that the structure corresponds to a support pole. We label the points in the cylindrical shell as pole points.

\subsubsection{Identifying tree membership of points (Tree Separation)}
\label{sec:seg}

\begin{figure*}[!ht]
\centering
\includegraphics[width=0.92\textwidth]{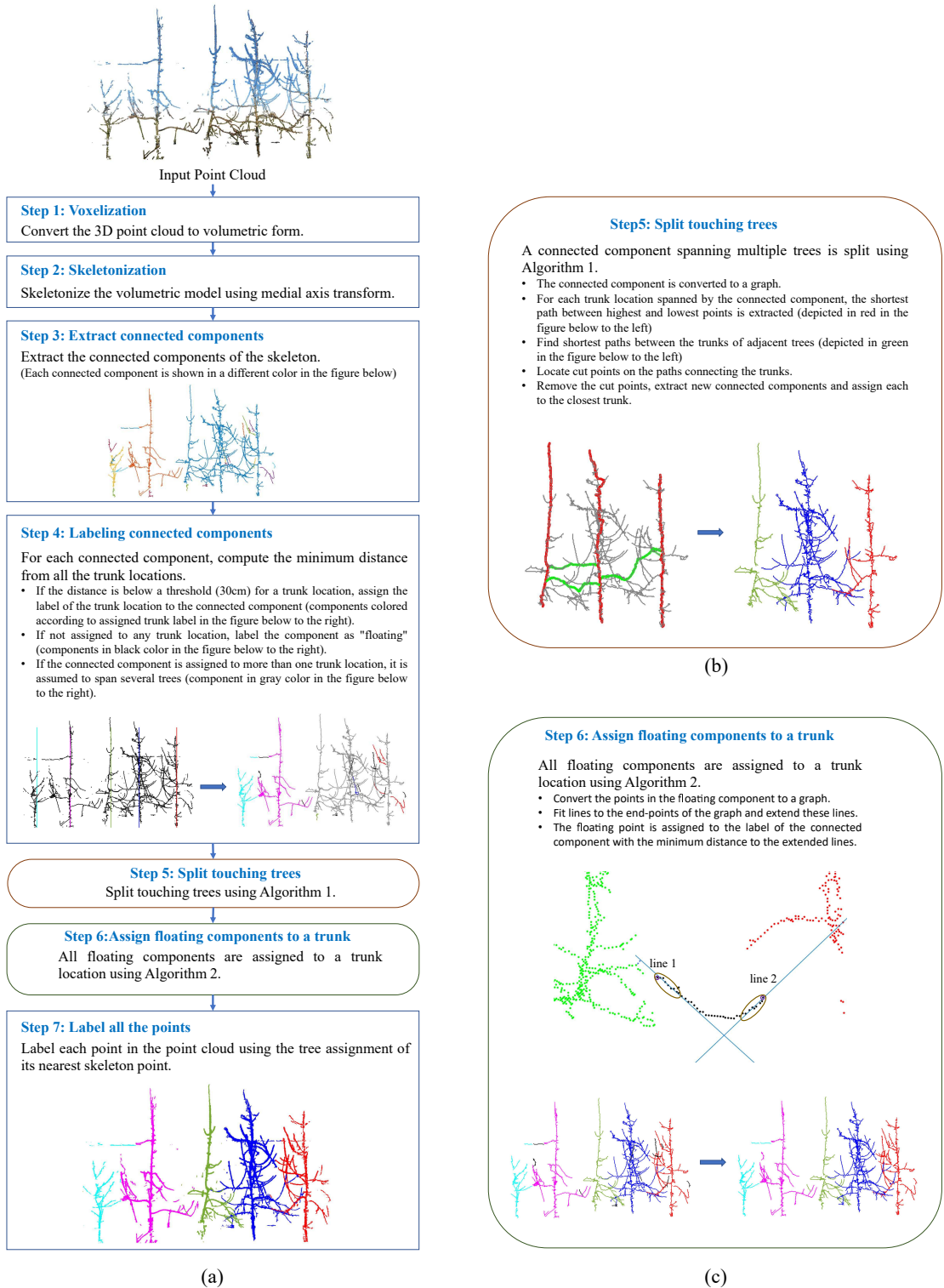}
\caption{(a) Block diagram for separating individual trees. (b) Illustration of Step 5 for splitting touching trees, (c) Illustration of Step 6 for labeling floating components.}
\label{fig:blockseparate}
\end{figure*}

This module of our pipeline is responsible for delineating the trees in $PC_{w}^{trees}$, which is the point cloud with trellis wires, the water-pipe and the support pole removed. The output of the delineation process is the assignment of each point in $PC_{w}^{trees}$ to one of the trees $T_j = (t_j,\hat{L}_j,SP_j) \in \mathcal{T}$.

The main steps of the tree separation process is given in Fig. \ref{fig:blockseparate}. We convert $PC_{w}^{trees}$ to binary volumetric form and apply skeletonization to conduct a connectivity analysis. We delineate adjacent trees if they are touching and we assign isolated connected components to one of the two nearest trees through a set of rules. The details of the steps are as follows:

\underline{\textit{Step 1: Voxelization}} The point cloud $PC_{w}^{trees}$ is converted to binary volumetric form $B_{trees}$ with voxel size $\hat{\Delta}_x = \hat{\Delta}_y =  \hat{\Delta}_z = 5mm$. 

\underline{\textit{Step 2: Skeletonization}} The skeleton $S_{trees}$ is extracted from $B_{trees}$ using medial axis thinning algorithm given in \cite{skeleton}. 

\underline{\textit{Step 3: Extraction of connected components}} Connected components of $S_{trees}$ are extracted using flood fill algorithm \cite{Shane2016}. We denote the set of connected components as $\mathcal{CC} = \{C_1,C_2,...,C_{N_{comp}}\}$, where $C_c$ is the $c$\textsuperscript{th} connected component and $N_{comp}$ is the number of connected components. Fig. \ref{fig:blockseparate} shows each connected component in a sample $S_{trees}$ in a different color.

\underline{\textit{Step 4: Labeling connected components}}
Using the trunk locations $\hat{L}_j = (0,\hat{y}_j,0)$, we compute the minimum distance of each connected component $C_c$ to all trunk locations. If this distance is below 30cm, then we assign $C_c$ to $t_j$. Fig. \ref{fig:blockseparate} (Step 4) gives the trunk locations as lines in different colors and the connected components colored according to the assigned tree for a sample $S_{trees}$. 

After this procedure a connected component might be assigned to 1) only one tree, 2) to multiple trees, or 3) none of the trees. If the connected component is assigned to multiple trees, it is assumed to be spanning several trees that are touching each other. We label the connected components not assigned to any tree as "floating". The floating components are shown in black color in Fig. \ref{fig:blockseparate} (Step 4). 

\underline{\textit{Step 5: Splitting touching trees}} For a connected component $C_c$ spanning $N_{c}$ trees $\{T_j\}$, $j = j_1^{c},...,j_{N_{c}}^c$, we run Algorithm \ref{alg:alg1}. Before running the algorithm, we update $SP_{j}$, \emph{main axis} of the $j$\textsuperscript{th} trunk, together with the points $\hat{p}_{j,{top}}$ and $\hat{p}_{j,{bottom}}$. Recall that $\hat{p}_{j,{top}}$ and $\hat{p}_{j,{bottom}}$ are the top and bottom points of the skeleton of the $j$\textsuperscript{th} tree trunk and $SP_{j}$ is the shortest path connecting them. Fig. \ref{fig:blockseparate} (b) shows a connected component spanning three trees. The main axes of them are plotted in red color, on the left.

\begin{center}
\begin{algorithm}[H]
 \KwData{$C_c$: Connected component spanning multiple trees; \\
 $\{T_j\}$: Trees spanned by $C_c$; $j = j_1^{c},...,j_{N_{c}}^c$; \\ 
 $\{SP_j\}$: Main axes of trees; \\
 $\{\hat{p}_{j,{top}}\}$: Top points of the main axes
}

 \KwResult{$\{C_{c,d}\}$: Detached connected components each assigned to a tree; $d = 1,...,{N_{comp}^{c}}$}
 
 \For{$j \gets j_1^{c}$ to $j_{N_{c}}^c-1$}{
    $CP \gets \emptyset$ \;
    Extract the shortest path $CP$ between the points $\hat{p}_{j,{top}}$ and $\hat{p}_{j+1,{top}}$\;
    \While{$CP \ne \emptyset$}{
    $CP \gets (CP \setminus SP_j) \setminus SP_{j+1}$ \;
    Select the global extremum of the z-coordinate in $CP$ as the cut-point\;
    Remove the cut-point from $C_c$  \;
    Extract the shortest path $CP$ between the points $\hat{p}_{j,{top}}$ and $\hat{p}_{j+1,{top}}$\;
   }
   }
 Apply connected components to $C_c$ to obtain $\{C_{c,d}\}$ \;
 Assign each connected component $C_{c,d}$ to the closest tree trunk\;
 \caption{Separation of a connected component into multiple trees}
 \label{alg:alg1}
\end{algorithm}
\end{center}

Algorithm \ref{alg:alg1} takes as input the set of trees identities $\{T_j\}$, $j = j_1^{c},...,j_{N_{c}}^c$ spanned by the connected component $C_c$. For each adjacent tree pair $T_j,T_{j+1}$, the shortest path between their top points $\hat{p}_{j,{top}}$ and $\hat{p}_{j+1,{top}}$ is extracted. We call this path $CP$, the connecting path, which contains the touching point of branches from trees $T_j$ and $T_{j+1}$. Each such path is searched for a cut-point to separate the connected adjacent trees. The cut-point is removed from the component $C_c$ to break the connectivity at that point. The process is repeated and $CP$ is updated until there remains no connected path between $\hat{p}_{j,{top}}$ and $\hat{p}_{j+1,{top}}$. Fig. \ref{fig:blockseparate} (b) depicts the connecting paths $CP$ between adjacent trees with green dots. 

After all connecting paths are extracted and the cut-points are found and removed, detached connected components $\{C_{c,d}\}$; $d = 1,...,{N_{comp}^{c}}$ of $C_c$ are extracted. Then each connected component is assigned to the tree identity of the closest tree trunk. Fig. \ref{fig:blockseparate} (b) shows the detached connected components each colored according to its tree identity.

It is challenging to determine the point where branches from two trees touch each other. Many architectural and morphological rules concerning apple tree branches can be incorporated. However, here, we use a simple heuristic based on the assumption that the point that changes direction along the z-axis (upwards or downwards) corresponds to a meeting point along the path. We select the global extremum of the z-coordinate as the cut-point of the connecting path.

\underline{\textit{Step 6: Assigning floating components to a tree}} The tree membership of a floating component $C_c$ is determined using Algorithm \ref{alg:alg2}. Before running Algorithm \ref{alg:alg2}, we identify the set $\mathbb{C}=\{(C_1,\tau_1),...,(C_{N_F},\tau_{N_F})\}$ of connected components already assigned to a tree. Here $\tau_f \in \{t_1,..t_{N_{trees}}\}$ is the tree identity of the component $C_f$. We determine the two closest components in $\mathbb{C}$ to the floating component $C_c$. If the distance to one connected component is more than 3 times than the distance to the other component, we assign the points in $C_c$ to the tree identity of the closest component. Otherwise, we locate the end-points in $C_c$, fit lines to these end-points and extend these lines, as shown in Fig. \ref{fig:blockseparate} (c). The minimum distance of the two closest connected components to these lines are calculated. The floating component is then assigned to the tree identity of the connected component with the minimum distance to the extended lines. Algorithm \ref{alg:alg2} gives the details of the process.

\underline{\textit{Step 7: Labeling all points with tree identities}} After Steps 5 and 6, all connected components in $S_{trees}$ are assigned to a tree label $\tau_c \in  \{t_1,..t_{N_{trees}}\}$. Recall that the connected components are extracted from the skeleton $S_{trees}$ of the point cloud $PC_{w}^{trees}$. For each point $\hat{p} \in PC_{w}^{trees}$, we locate the closest component of $S_{trees}$ and assign the tree identity of the component to the point $\hat{p}$. Fig. \ref{fig:blockseparate} (Step 7) shows the points of a sample $PC_{w}^{trees}$ colored according to their tree identities.

\begin{center}
\begin{algorithm}[H]

 \KwData{$C_c$: Floating connected component;\\ $\mathbb{C}=\{(C_f,\tau_f)\}$: Connected components already assigned to a tree ($f = 1,2,...,N_F$) }
 
 \KwResult{$\tau_c \in  \{t_1,..t_{N_{trees}}\}$: Tree identity of $C_c$ }
 \For{$f \gets 1$ to $N_F$}{
 Calculate the minimum distance $d_f$ between the points in $C_c$ and the points in $C_f$\;
   }
$d^{F1} \gets$ Minimum of $d_f$;
$\quad d^{F2} \gets$ Next minimum of $d_f$\;
$C^{F1} \gets$ Component with $d^{F1}$ ;
$\quad C^{F2} \gets$ Component with $d^{F2}$\;
$\tau^{F1} \gets$ Tree identity of $C^{F1}$ ;
$\quad \tau^{F2} \gets$ Tree identity of $C^{F2}$\;
\eIf{$\frac{d^{F2}}{d^{F1}} > 3$}{
$\tau_c \gets \tau^{F1}$\;
}
{
Extract the end-points $p_e$ of $C_c$, $e=1,2,...,N_e$\;
\For{$e \gets 1$ to $N_e$}{
 Extract $K$ nearest neighbors of $p_e$ with $K = 10$ \;
 Fit a line $l_e$ to the neighbors\;
 $d^{eF1} \gets$ Minimum distance of the points in $C^{F1}$ to the line $l_e$\;
  $d^{eF2} \gets$ Minimum distance of the points in $C^{F2}$ to the line $l_e$\;
   }
\eIf{$min\{d^{eN1}\}< min\{d^{eN2}\} $}{
$\tau_c \gets \tau^{F1}$\
} {$\tau_c \gets \tau^{F2}$\;} 
}

 \caption{Assignment of a floating branch to a neighboring tree.}
 \label{alg:alg2}
\end{algorithm}
\end{center}

Recall that $PC_{w}^{trees}$ is a subset of $PC_{w}^{TP}$, which is the winter point cloud aligned to the trellis-plane. To find the tree identities of the points in the calibrated winter cloud $PC_{w}^{C}$, we first apply $p = \hat{p}R^{-1}$ to each point $\hat{p} \in PC_{w}^{trees}$ with tree identity $\tau \in \{t_1,...,t_{N_{trees}}\}$. Then, we retrieve the closest point $p_i \in PC_{w}^{C}$ to $p$ and set $\tau_i = \tau$.

\subsection{Apple detection}

To detect apples, we applied simple color thresholding to the calibrated 3D color point cloud of the harvest scene $PC_{h}^{C}$. First, the RGB colors of points are converted to HSV (Hue, Saturation, Value) representation. The points in the hue range [0.15-0.2] are assumed to correspond to green/yellow apple points. The red apple points are assumed to be in the hue range [0-0.05] and [0.95-1]. The points with hue values in these ranges are retrieved and converted to volumetric form. The connected components of the volumetric form and their bounding boxes are extracted. The centers of these bounding boxes are mapped to the 3D space of $PC_{h}^{C}$ and are considered to be the locations of detected apples. We denote the set of detected apples in a harvest scene as $\mathcal{A} = \{p_1^{\alpha},...,p_{N_{apples}}^{\alpha}\}$, where $p_a^{\alpha}$ is the location of a detected apple.

Although our apple detection approach is primitive, it provides recall rates in the range of 74\% to 90\% (see Section \ref{sec:resultapples}). This level of detection success is sufficient to demonstrate the effectiveness of our approach for assigning retrieved apples to their respective trees.

\subsection{Assigning apples to individual trees}
\label{sec:assignment}

The main objective of this work is to automatically assign detected apples to their respective trees; i.e. to determine the tree identity $\tau_a \in \{t_1,...,t_{N_{trees}}\}$ of each detected apple $p_a^{\alpha} \in \mathcal{A}$. To this end, we align calibrated winter cloud $PC_{w}^{C}$ and summer cloud $PC_{h}^{C}$ and assign apple $p_a^{\alpha}$ detected from $PC_{h}^{C}$ to the tree identity of the closest branch point in the aligned winter cloud.

Since both point clouds were transformed, through calibration, to a common reference frame with the origin at the base of a reference tree (see Supplementary Material A for details), they are already initially aligned. We apply the standard Iterative Closest Point (ICP) algorithm \cite{chen1992object} to improve the alignment. Point to point metric is used to minimize the alignment error. ICP returns the transformation parameters; a rotation matrix $R^{wh}$ and a translation vector $T^{wh}$ that align the points in $PC_{w}^{C}$ to the points in $PC_{h}^{C}$:
\begin{equation}
    PC_{wh}^C = \{p_i' = p_i R^{wh} + T^{wh} : (p_i \in PC_{w}^{C}) \text{ } \& \text{ } \tau_i \in \{t_1,...,t_{N_{trees}}\} \}.
\end{equation}

Once the transformed winter point cloud $PC_{wh}^C$ is obtained, the closest branch point in $PC_{wh}^C$ labeled with a tree identity to the apple location $p_a^{\alpha}$ is retrieved:
\begin{equation}\label{eq:appleid1}
    i^* = \arg \min_{ p_i' \in PC_{wh}^C}  \| p_i' - p_a^{\alpha} \|; 
\end{equation}

\noindent and the tree identity of apple $a$ is set as
\begin{equation}\label{eq:appleid2}
    \tau_a = \tau_{i^*}.
\end{equation}

\subsection{Ground truth and evaluation metrics}

\begin{figure*}[!ht]
\centering
\includegraphics[width=0.43\textwidth]{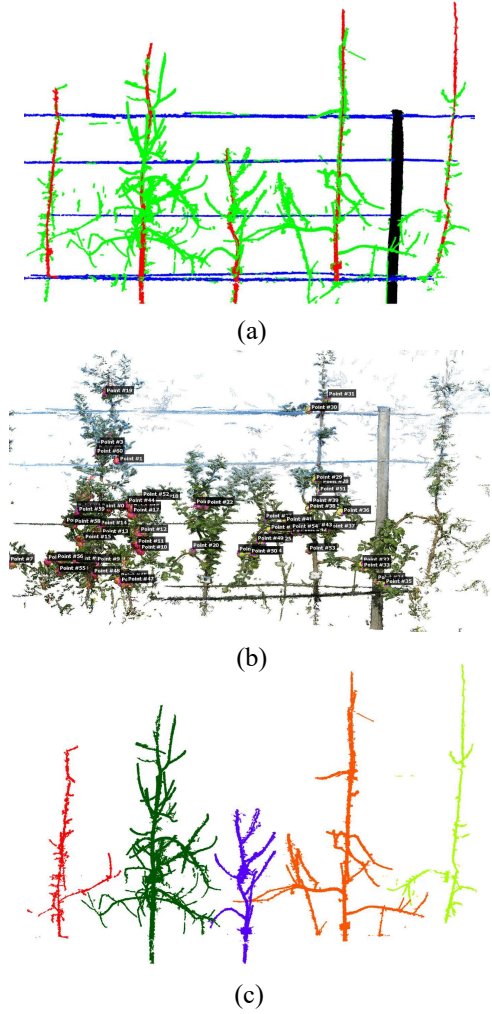}
\caption{Ground truth. (a) Manually labeled point cloud for assessment of trellis wire, tree trunk and support pole detection, (b) Harvest point cloud with ground truth apple locations, (c) Point cloud manually segmented to individual trees.}
\label{fig:GT}
\end{figure*}

To provide ground truth for evaluation of our semantic segmentation scheme, we manually labeled each point  $p_i \in PC_{w}^{C}$ with one of the semantic labels $\gamma_i^{GT} \in \{\textit{"Tree trunk"}, \textit{"Branch"}, \textit{"Trellis wire+Water pipe"}, \textit{"Support pole"}\}$. We used CloudCompare (2.11, GPL software, 2020) to label the point cloud. Fig. \ref{fig:GT}-(a) shows a sample winter scene with points colored according to their manually annotated ground truth labels.
 
We evaluated the performance of the semantic segmentation module described in Section \ref{sec:trellis} using Recall ($Re$), Precision ($Pr$), F1 score ($F1$), Intersection over Union ($IoU$), and Class Accuracy ($CA$), defined as

\begin{equation}\label{eq:recall}
Re = \frac{TP}{TP+FN}
\end{equation}

\begin{equation}\label{eq:precision}
Pr = \frac{TP}{TP+FP}
\end{equation}

\begin{equation}\label{eq:F1-score}
F1= 2\times\frac{Pr \times Re}{Pr + Re} 
\end{equation}

\begin{equation}\label{eq:IoU}
 IoU = \frac{TP}{TP+\;FN+\;FP} 
\end{equation}

\begin{equation}\label{eq:class accuracy}
CA = \frac{TP+TN}{TP+TN+FP+FN}\;,
\end{equation}

\noindent where $TP$, $TN$, $FP$ and $FN$, correspond to the number of True Positives, True Negatives, False Positives, and False Negatives, respectively. These cases for the \textit{"Tree trunk"} are determined as follows:
\begin{equation}
\text{Case}_i = 
\begin{cases}
     \text{True Positive} & \text{if } \gamma_i = \gamma_i^{GT} = \text{\textit{"Tree trunk"}}
     \\ 
      \text{True Negative} & \text{if } (\gamma_i \neq \text{\textit{"Tree trunk"}}) \& (\gamma_i^{GT} \neq \text{\textit{"Tree trunk"}})
     \\ 
     \text{False Positive} & \text{if } (\gamma_i = \text{\textit{"Tree trunk"}}) \& (\gamma_i^{GT} \neq \text{\textit{"Tree trunk"}})
     \\ 
     \text{False Negative} & \text{if } (\gamma_i \neq \text{\textit{"Tree trunk"}}) \& (\gamma_i^{GT} = \text{\textit{"Tree trunk"}}),
\end{cases}
\end{equation}

\noindent where $\gamma_i^{GT}$ is the ground truth label of point $p_i$ and $\gamma_i$ is the label predicted by our automatic semantic segmentation scheme. The cases for \textit{"Trellis wire+Water pipe"} and \textit{"Support pole"} are obtained in a similar manner.

In order to assess the performance of the color-based apple detection approach, we manually marked the apple positions in the harvest point clouds and obtained the set of points $\mathcal{A}^{GT} = \{p_g^{\alpha,GT}\}; g = 1,...,N_{apples}^{GT}$. In Fig. \ref{fig:GT}-(b), a harvest point cloud with ground truth apple positions is shown. For evaluation, we used Recall ($Re$) and Precision ($Pr$) metrics, defined in Eq. (\ref{eq:recall}) and (\ref{eq:precision}). Here, the True Positives correspond to the cases where a ground truth apple is correctly localized. The False Positives are wrong detections returned by the algorithm. The False Negatives correspond to the ground truth apple locations missed by the algorithm. A detection $p_a^{\alpha} \in \mathcal{A}$ is considered a True Positive if there is a ground truth apple $p_g^{\alpha,GT} \in \mathcal{A}^{GT}$ such that $\| p_a^{\alpha}-p^{\alpha,GT} \|<10cm$ and there is no other detected apples closer to $p^{\alpha,GT}$. We pair the indices $(a,g)$ to indicate that $p_a^{\alpha} \in \mathcal{A}$ corresponds to $p_g^{\alpha,GT} \in \mathcal{A}^{GT}$. The number of False Positives and False Negatives are then calculated as:
\begin{equation}
    FP = N_{apples} - TP
\end{equation}
\begin{equation}
    FN = N_{apples}^{GT} - TP
\end{equation}
\noindent where $TP$ is the number of True Positives, $N_{apples}$ is the number of detected apples in $\mathcal{A}$ and $N_{apples}^{GT}$ is the number of ground truth apples in $\mathcal{A}^{GT}$.

The end result of our apple assignment pipeline is the tree identity of each detected apple, indicating which tree it belongs to. In order to evaluate assignment performance, we provided the correct tree identities of the ground truth apples via manual inspection; i.e. we determined $\tau_g \in \{1,..,N_{trees} \}$ for each $p_g^{\alpha,GT} \in \mathcal{A}^{GT}$. We computed the accuracy of the apple assignment ($ACC$) as the ratio of the number of correctly assigned true positives $TP_C$ to the total number of true positives $TP$ in the scene:

\begin{equation}
    ACC = \frac{TP_C}{TP}
\end{equation}

A detection $p_a^{\alpha} \in \mathcal{A}$ is considered to be a correctly assigned true positive if its tree identity $\tau_a$, determined by Eq. (\ref{eq:appleid1}) and (\ref{eq:appleid2}), is equal to the tree identity $\tau_g$ of its matched ground truth apple $p_g^{\alpha,GT} \in \mathcal{A}^{GT}$.

 Recall that we assigned each apple $p_a^{\alpha} \in \mathcal{A}$ to the tree identity $\tau_{i^*}$ of the closest branch point $p_{i^*}$ in the aligned winter cloud through Eq. (\ref{eq:appleid1}) and (\ref{eq:appleid2}). In order to decouple the apple assignment errors due to branch deformation between winter and summer trees and errors due to our automatic tree separation method, we performed the apple assignment procedure on two types of data: 
 \begin{enumerate}
 \item \underline{Manually Separated:} We manually separated the winter point clouds into individual trees and provided the ground truth tree identities $\tau_{i}^{GT} \in \{1,...,N_{trees}^{GT} \}$ of the trunk and branch points in the winter cloud. We used CloudCompare (2.11, GPL software, 2020) for annotation. One example is shown in Fig. \ref{fig:GT}-(c).
 \item \underline{Automatically Separated:} We used the tree identities $\tau_{i} \in \{1,...,N_{trees}\}$ of the trunk and branch points in the winter cloud predicted by our automatic tree separation procedure.
 \end{enumerate}


\section{Results}

We first report the results of the semantic segmentation method, which detects the trellis wires, tree trunks and support poles. Then, we provide the performance of the apple detection method and the assignment procedure of apples to individual trees in the scene.

\begin{figure}[!ht]
\centering

\begin{tabular}{{ccc}}

\includegraphics[width=0.26\textwidth]{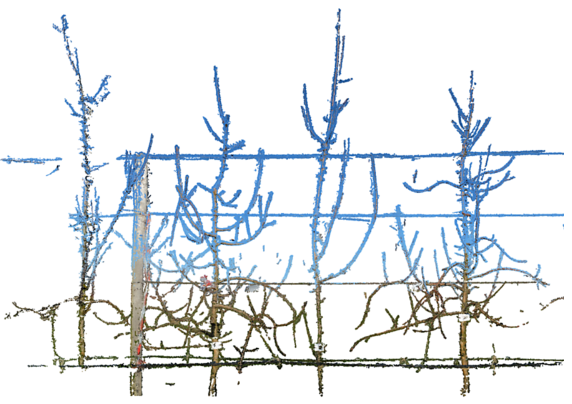} & \includegraphics[width=0.26\textwidth]{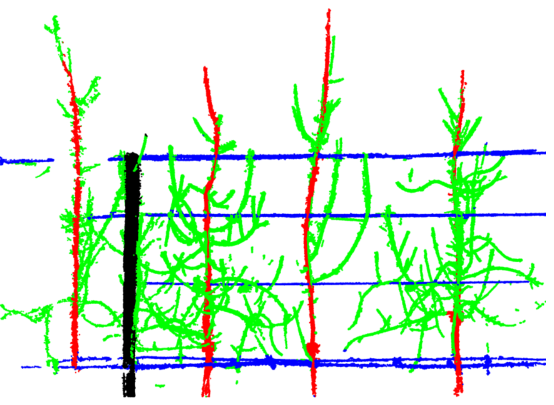}  & \includegraphics[width=0.26\textwidth]{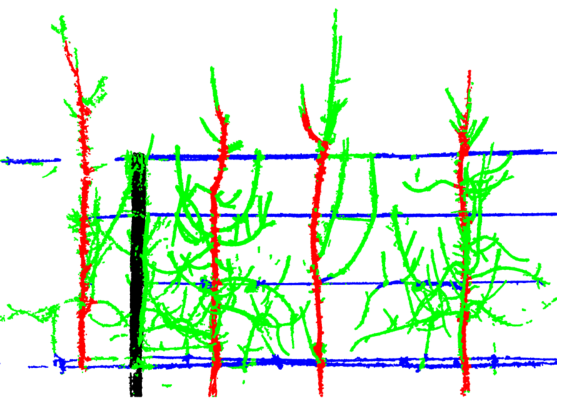} \\

\includegraphics[width=0.28\textwidth]{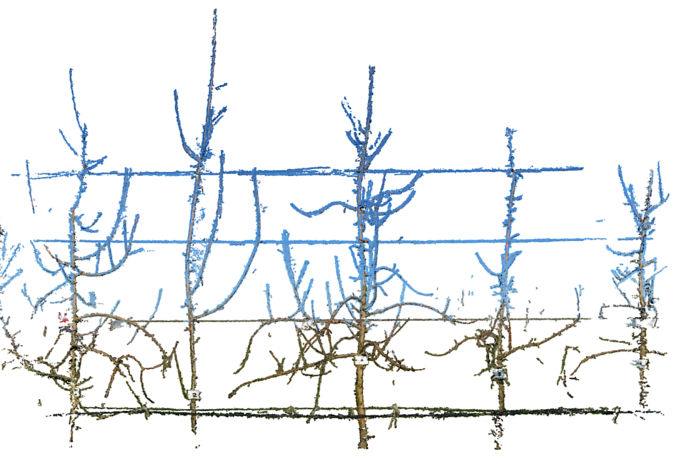} & \includegraphics[width=0.28\textwidth]{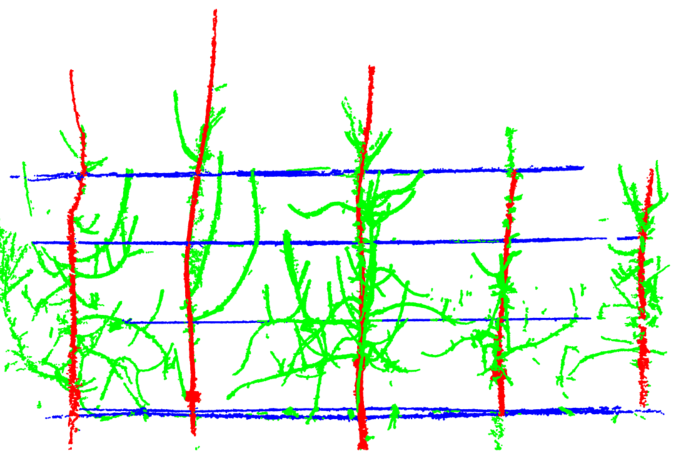}  & \includegraphics[width=0.28\textwidth]{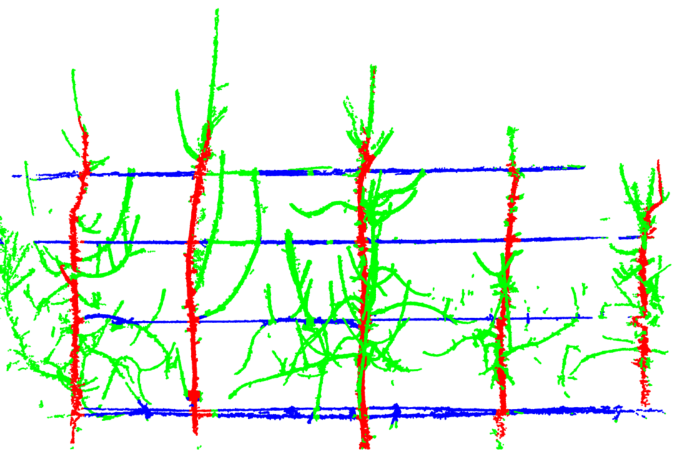} \\
(a)  & (b) & (c) \\
\end{tabular}
\caption{(a) Calibrated point clouds, (b) Manually generated Ground Truth (cyan:trellis wires, red: tree trunks, black: support poles), (c) Semantic labels obtained by our method for automatic detection of trellis wires, tree trunks, and support poles}
\label{table:Numofimages} 

\end{figure}

\begin{table}[!ht]
    \centering
    \caption{Performance of the method for detection of trellis wires, tree trunks and support poles. NP is for non-present.}
    \begin{tabular}{{cccccc}}
        \hline
        \multirow{2}{*}{}&\multicolumn{5}{c}{\textbf{Trellis wires}}\\\cline{2-6}
         & \% $Re$ & \% $Pr$ & \% $F1$ &\% $IoU$ & \% $CA$ \\\hline
        
        Scene 1 & 84.98 &	81.61 &	83.26 &	71.32 &	96.42 \\\hline
        Scene 2 & 88.16 &	76.01 &	81.63 &	68.96 &	95.19 \\\hline
        Scene 3 & 91.48	&   73.65 &	81.61 &	68.93 &	95.52 \\\hline
        Scene 4 & 86.48	&   88.20 &	87.33 &	77.51 &	96.88 \\\hline
        Scene 5 & 75.47	&   85.21 &	80.04 &	66.73 &	95.64 \\\hline
        Scene 6 & 85.82 &	77.75 &	81.59 &	68.90 &	96.06 \\\hline
        Scene 7 & 79.24 &	81.16 &	80.19 &	66.93 &	96.48 \\\hline
       
        &\multicolumn{5}{c}{\textbf{Tree trunks}}\\\cline{2-6}
        & \% $Re$ & \% $Pr$ & \% $F1$ &\% $IoU$ &\% $CA$ \\\hline
        
        Scene 1 & 90.26 & 77.97 & 83.67 & 71.92 & 92.83 \\\hline
        Scene 2 & 91.49 & 74.89 & 82.36 & 70.01 & 91.77 \\\hline
        Scene 3 & 92.77 & 70.03	& 79.81 & 66.40 & 91.31 \\\hline
        Scene 4 & 83.23 & 71.23 & 76.76 & 62.29 & 90.55 \\\hline
        Scene 5 & 94.25	& 67.03 & 78.34 & 64.40 & 91.56 \\\hline
        Scene 6 & 94.02 & 70.39 & 80.51 & 67.37 & 92.78 \\\hline
        Scene 7 & 95.47 & 69.19 & 80.24  & 66.99 & 93.27 \\\hline

        &\multicolumn{5}{c}{\textbf{Support poles}}\\\cline{2-6}
        & \% $Re$ & \% $Pr$ & \% $F1$ &\% $IoU$ &\% $CA$ \\\hline
        
        Scene 1 & 95.50 & 96.24 & 95.87 & 92.07 & 99.44 \\\hline
        Scene 2 &  NP     & NP &  NP & NP  &NP \\\hline
        Scene 3 &  NP     & NP &  NP & NP  &NP \\\hline
        Scene 4 &  NP     & NP &  NP & NP  &NP\\\hline
        Scene 5 & 91.83 & 98.74 & 95.16 & 90.77 & 98.65 \\\hline
        Scene 6 & 94.44 & 99.26 & 96.79 & 93.78 & 98.15 \\\hline
        Scene 7 & 97.94 & 98.96 & 98.45 & 96.94 & 99.64 \\\hline
    \end{tabular}
\label{table:object detection assessement} 
\end{table}

\subsection{Evaluation of detection of trellis wires, tree trunks and support poles}

In Fig. \ref{table:Numofimages}, we give visual results of our semantic segmentation method for two winter scenes. The visual results for all the seven scenes can be found in Supplementary Material B. We can observe that all the trees in the scenes of the apple orchard, the trees were correctly localized. The number of detected tree trunks and the actual number of trees were equal for all seven scenes; $N_{trees} = N_{trees}^{GT}$. 

Table \ref{table:object detection assessement} provides quantitative evaluation of our semantic segmentation method. The recall and precision values for the trellis wires are satisfactory. All the support poles in the scenes were correctly identified and segmented with over 90\% success. The recall rate for the trunks is over 90\% for all but one scene, meaning that most of the trunk points are retrieved. The precision rates are satisfactory for our purposes. The less than perfect precision is due to the fact that branching points close to the tree trunks are also classified as trunks by our method. 

It should be recalled that our aim is not to provide a perfect segmentation, but rather 1) to detect and remove the trellis wires to break connectivity between adjacent trees, 2) to locate the tree trunks correctly to be able to separate individual trees, and 3) to remove the support poles. For the purposes of our application, these aims were achieved with this level of automatic point labeling of the scene.

\subsection{Evaluation of apple detection and assignment to individual trees}
\label{sec:resultapples}
\begin{figure*}[!ht]
\centering
\includegraphics[width=0.8\textwidth]{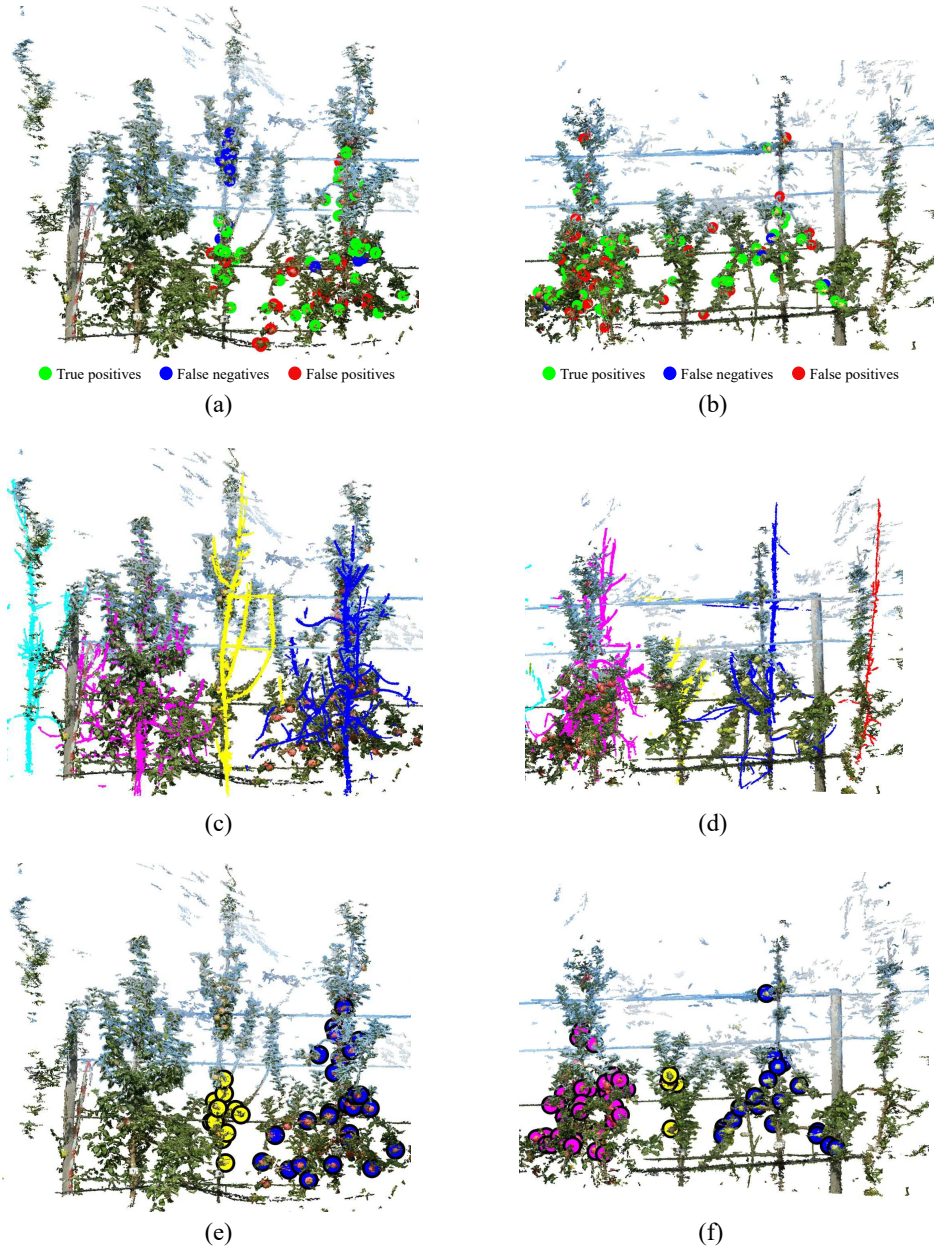}
\caption{(a), (b) True positives, false negatives, and false positives obtained with color-based apple detection method for two sample scenes. (c), (d) Registration of harvest and winter clouds for the two scenes. Each separated tree is shown with a different color. (e), (f) Assignment of true positives to their corresponding trees for the two scenes.}
\label{fig:assignmentdem}
\end{figure*}

The precision and recall values obtained with color-based apple detection are given in Table~\ref{table:Apple detection assessment}. 
Despite the simplicity of the detection approach, we achieved over 90.75\%  recall; i.e. most of the apples in the ground truth were retrieved. The false negatives occurred since we did not post-process the connected components for resolving clusters of apples. The over-detection (precision 65,37\%) can be explained by the sensitivity of the color-based algorithm and the lack of shape-based apple verification. Fig. \ref{fig:assignmentdem} (a) and (b) visually illustrate the performance of our apple detection method on two sample scenes. 

\begin{figure*}[!ht]
\centering
\includegraphics[width=0.8\textwidth]{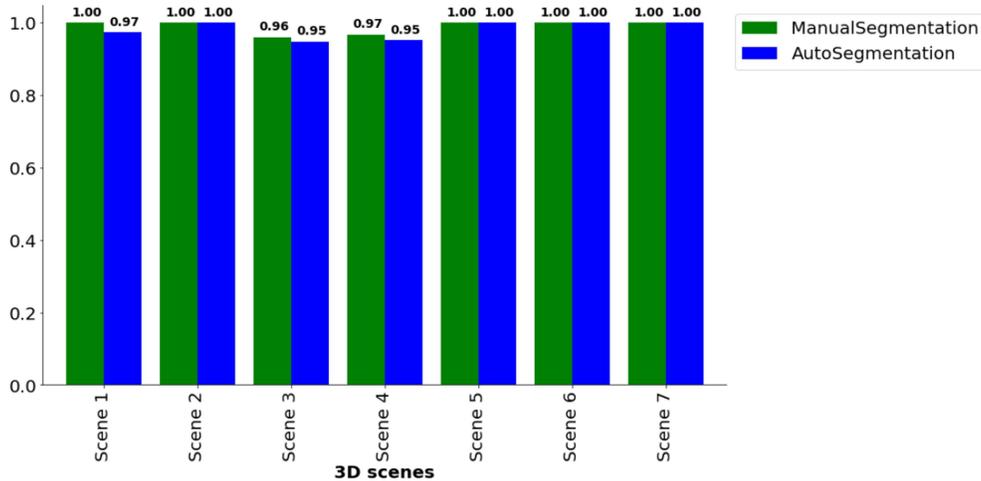}
\caption{Accuracy of assigning apples to the correct apple trees in 3D models.}
\label{accyracy}
\end{figure*}

Our main task is to correctly assign the detected apples to the individual trees they belong to. 
As we have stated earlier, we performed the assignment procedure to two types of data: 
1) The winter point clouds which are manually segmented to individual trees, and 2) The winter point clouds where the trees are segmented using our automatic tree separation method. Fig. \ref{accyracy} shows the assignment accuracy ($ACC$) on both type of data. The performance is high for both cases (100\% on four scenes). With automatic tree separation, a performance drop of less than 3\% is observed, demonstrating that our automatic pipeline was able to detach individual trees and correctly assign the detected apples.

\begin{table}[!ht]
\centering
\caption{Apple detection performance.}
\label{table:Apple detection assessment} 

\begin{tabular}{{ccc}}
\hline
3D scenes& \% $Re$ & \% $Pr$ \\\hline
Scene 1 & 74.50  & 61.29 \\\hline
Scene 2  & 87.34 &  62.16\\\hline
Scene 3  &  88.54 &  58.21 \\\hline
Scene 4  & 90.00 & 48.64 \\\hline
Scene 5  & 90.62 & 58.58 \\\hline
Scene 6  &  77.41 & 65.62 \\\hline
Scene 7  & 80.85 & 66.66 \\\hline
\end{tabular}
\end{table}

\bigskip

Fig. \ref{fig:assignmentdem} (c) and (d) show the registration result of winter and harvest point clouds for two sample scenes. Each separated tree in the winter clouds is shown in a different color. In Fig. \ref{fig:assignmentdem} (e) and (f), the detected apples are shown with the color of their corresponding tree labels. 

\section{Discussion}

The full pipeline presented and tested in this manuscript achieves great performance for assigning apples to individual trees in dense orchards. The main strategy is aligning summer and winter point clouds. The sub-steps of the pipeline, for which we chose standard approaches for implementation, are open to improvement for further performance increase. 

Images were acquired manually with a standard camera. This is a rather time consuming process for producing hundreds of images per tree. The speed of acquisition can be increased and the amount of images can be optimized by a drone with a camera or a land robot with multiple cameras and automatic navigation via GPS localization \cite{mogili2018review}. The object of reference for calibration and registration of the summer and winter point cloud was chosen to be the X-Rite ColorChecker, since it is a standard tool in the computer vision community. In principle, any reference object with a distinctive geometric pattern could serve the same purpose. One could, for instance, use a large data matrix mounted on a tripod stick. The data matrix could enable both tree identification and calibration of the acquired 3D data.

The deformation we observed with our data (young trees of four years old) becomes even more pronounced for older trees. Registration of winter and summer calibrated point clouds could be performed efficiently with non-rigid registration while dealing with older trees, where the deformation during summer could be larger due to increased fruit load. Non-rigid registration is widely used in medical imaging when data from two different modalities, such as MRI and X-Ray images, should be registered. Non-rigid deformation between the image sets are commonly observed due to movement of the patient or artifacts of the imaging systems. The literature on non-rigid registration of medical images can thus be revisited for our plant imaging problem \cite{holden2007review}. To avoid having a too large exploration space for this non-rigid registration, one could also use botanical and physical knowledge on the development of trees. The size and weight of the fruits is important because it can cause arching of the branches, therefore, a deformation of the architecture. Another factor that alters the architecture is the secondary growth of the branches. Expert knowledge on such processes can be used to constrain the deformation space and fix the hyperparameters of the non-rigid registration algorithms. 

In this work, we used connectivity analysis and simple heuristics to disconnect touching trees. Alternatively, the identification of each tree unit can be achieved using the architectural criteria specific to each tree. They are linked to the basic architectural models defined for each taxon \cite{Halle2004}. They are supplemented by the growth conditions specific to each tree and are assessed by the diameter, length, age and branching angles of the branches but also by the location of inflorescences and fruits. 

Last but not least, the apple detection algorithm chosen in this manuscript was extremely simple and it will be necessary to revisit the huge literature on apple detection to improve the performance, specially on groups of apples or to reduce the amount of false positives. State-of-the-art methods employing deep learning architectures, such as \cite{roy2019vision, hani2018apple, hani2020comparative}, can be employed for highly accurate apple detection and counting.

Our pipeline enables the assignment of apples to the trees that bear them. This makes it possible to assess the production and the quality of the fruiting body in variety testing applications and also in the agronomic management of orchards. We know that fruiting is the expression of primary and secondary growth followed by a flowering process with the formation of inflorescences and flowers. One, two or three years old axes that are part of the overall architecture of the tree carry these inflorescences. In this biological process, Laury et al. \cite{Lauri1996,Lauri2010} showed the importance of the age of branches, their position in the architecture and secondary growth on the fruit load of the tree. Our pipeline opens the way to acquire data at different developmental stages, analyze the architecture of individual trees, track primary and secondary growth, determine their axes of different ages. The location of the fruits and the identification of the characteristics of the axes that carry them, supplemented by a temporal monitoring of the architectural development could make it possible to obtain information to manage and improve the agronomic management of fruit trees.

\section{Conclusion}

In this article, we presented, for the first time to the best of our knowledge, a pipeline to assign detected apples to their corresponding apple trees in 3D color point clouds. The pipeline was able to detect and filter out trellis wires and support poles. It successfully located trunk locations in the scene and retrieved trunk points with more than 90\% recall rate. The detected apples were assigned to their corresponding trees with more than 95\% accuracy.

This first proof of feasibility has shown the possibility and benefit of registration of 3D models of orchard scenes obtained in two different seasons. A direction for further development could be more frequent acquisition and reconstruction during the year, for instance during flowering period to link flower density to apple yield on individual trees. As another application, the configuration of the fruits in the harvest period can be used to guide the pruning process in early spring. These perspectives are now open with the pipeline proposed in this study. 

\bibliographystyle{elsarticle-num-names}

\bibliography{SeparationAppleTreesArXivVersion.bib}

\end{document}